\pdfoutput=1

\documentclass[11pt]{article}

\usepackage[preprint]{acl}

\usepackage{times}
\usepackage{latexsym}
\usepackage[T1]{fontenc}
\usepackage[utf8]{inputenc}
\usepackage{microtype}
\usepackage{amsmath}
\usepackage{graphicx}
\usepackage{multicol}
\usepackage{adjustbox}
\usepackage{booktabs}
\usepackage{multirow}
\usepackage{float}
\usepackage{cleveref}
\usepackage{caption}
\usepackage{subcaption}
\usepackage{amssymb}
\usepackage[normalem]{ulem}
\useunder{\uline}{\ul}{}
\usepackage{wrapfig}

\usepackage[T1]{fontenc}

\usepackage[utf8]{inputenc}

\usepackage{microtype}

\usepackage{inconsolata}

\usepackage{xspace}
\newcommand{\ie}{\emph{i.e.}\xspace}
\newcommand{\eg}{\emph{e.g.}\xspace}
\newcommand{\ia}{\emph{i.a.}\xspace}
\newcommand{\wrt}{wrt\xspace}

\newcommand{\probar}{\textsc{ProbAR}\xspace}

\title{Variability Need Not Imply Error: \\ The Case of Adequate but Semantically Distinct Responses}

\author{Evgenia Ilia \\
  University of Amsterdam \\
  \texttt{e.ilia@uva.nl} \\\And
  Wilker Aziz \\
  University of Amsterdam \\
  \texttt{w.aziz@uva.nl} \\}

\begin{document}
\maketitle
\begin{abstract}

With the broader use of language models (LMs) e need to  estimate their ability to respond reliably to prompts (\eg, are generated responses likely to be correct?). 
Uncertainty quantification tools (notions of confidence and entropy, \ia) 
can be used to that end 
(\eg, to reject a response when the model is `uncertain').
For example, \citet[semantic entropy;][]{kuhn2022semantic} regard \emph{semantic} variation amongst sampled responses 
as evidence that the model `struggles' with the prompt and that the LM is likely to err. 
We argue that semantic variability need not imply error---
this being especially intuitive in open-ended settings, where prompts  elicit multiple \emph{adequate} but semantically distinct responses.
Hence, we propose to annotate sampled responses for their adequacy to the prompt (\eg, using a  classifier) and estimate the 
\underline{Prob}ability the model assigns to \underline{A}dequate \underline{R}esponses (\probar), which we then regard
as an indicator of the model's reliability at the instance level.
We evaluate \probar as a measure of confidence in selective prediction with OPT models
(in two QA datasets and in next-word prediction, for English) and  
 find \probar to outperform semantic entropy 
across prompts with varying degrees of ambiguity/open-endedness. 

\end{abstract}

\section{Introduction}

\begin{figure}[t!]
    \centering
        \includegraphics[width=7.5 cm]{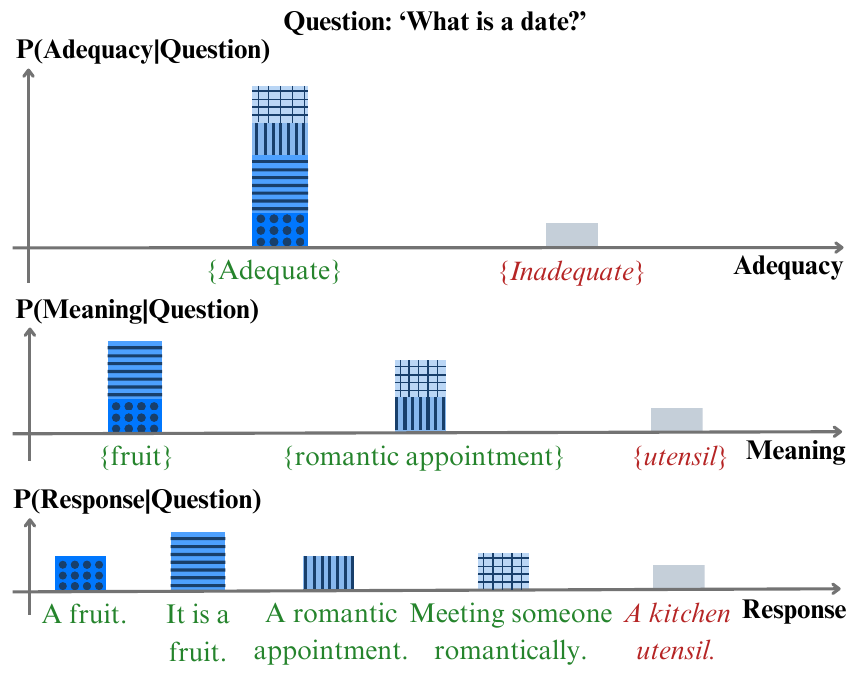}
        \caption{Bottom: a sampled-based approximation of an LM's distribution over responses given the question `\texttt{What is a date?}'; while the distribution exhibits high entropy, some answers are semantically equivalent. Middle: responses are clustered by meaning; 
        while this representation still exhibits high `semantic entropy' \citep{kuhn2022semantic}, probability concentrates on answers to different but plausible interpretations of the question. Top: responses are grouped as a function of their adequacy 
        to the prompt (\ie, \wrt any of its plausible interpretations); we regard the  probability accumulated by adequate responses ({\small \probar}) as an expression of confidence and expect it to  predict a model's instance-level performance on both more and less ambiguous/open-ended prompts. 
        }  
        \label{fig:space_repart}
\end{figure}

The use of LMs as tools to aid in decision-making 
evokes the need for techniques that help us determine when they can reliably respond to a prompt \citep{kuhn2022semantic, lin2024generating, gruber2024bias}. 
Indicators of uncertainty in generation, \eg based on LMs' predictive distributions \citep{fomicheva-etal-2020-unsupervised, aina-linzen-2021-language}  or statistics of sampled responses \citep{ren2022out, lin2024generating, manakul-etal-2023-selfcheckgpt, van-der-poel-etal-2022-mutual},
have been shown useful to that end, with LMs exhibiting higher uncertainty when they respond incorrectly.
These uncertainty quantifiers enable \emph{selective prediction} \citep[\ie, accepting/rejecting a response depending on the LM's degree of uncertainty;][]{kamath-etal-2020-selective, varshney-etal-2022-investigating}.
For example, entropy (a quantification of the amount of variation in sampled responses)
has been shown to (anti)-correlate with generation quality \citep{fomicheva-etal-2020-unsupervised, xu-etal-2020-understanding-neural} and a good predictor of error  \citep{malininuncertainty, xiao-wang-2021-hallucination, rawte-etal-2023-troubling}. 
For short generations, \citet{kuhn2022semantic} associate variation in meaning 
with propensity for error, showing their semantic entropy 
to outperform entropy's potential for selective prediction.

In many applications (\eg, question answering, dialogue, story generation), due to factors such as ambiguity, underspecification and differences in perspectives or beliefs \citep{plank-2022-problem, jiang-marneffe-2022-investigating,aroyo2015truth,baan2023uncertainty}, 
 prompts may elicit responses that differ widely while still being arguably adequate. 
In such cases, which we  refer to as `open-ended', 
the mere presence of semantic variation amongst sampled responses says little, if anything, about an LM's ability to respond correctly.
For a more generally applicable uncertainty quantifier, robust to prompts of varying open-endedness, we argue that rather than semantic similarity amongst sampled responses, we need to reason about whether or not sampled responses are generally \emph{adequate} to the prompt---see \Cref{{fig:space_repart}}. 
We propose to estimate the \underline{Prob}ability that an LM assigns to \underline{A}dequate \underline{R}esponses (\probar) and regard that as a notion of confidence in the LM's ability to respond to a prompt. 
For that, we design adequacy classifiers to approximately determine the rate at which sampled responses are independently judged to be adequate. 

We experiment with three English datasets \citep[][Abg-COQA, AmbigQA, Provo, resp.]{guo2021abg,min-etal-2020-ambigqa,luke2018provo} containing prompts of varying open-endedness, and use \probar as a measure of confidence in selective prediction with OPT models \citep{zhang2022opt}. We compare \probar to entropy, semantic entropy and a variant of P(True) \citep{kadavath2022language}. 
We find \probar to outperform them (in terms of AUROC and selective precision vs. coverage), with this being true for both more and less ambiguous/open-ended prompts.
For a subset of Abg-COQA, we estimate upperbounds on the  performances of semantic entropy and \probar by replacing their automated components (clustering algorithm and adequacy classifier, resp.) by human judgement. 
The Appendix reports extensively on the development of the adequacy classifiers 
(including human evaluation) as well as on factors such as sample size (for \probar estimation), decoding strategy (for selective prediction) and experimental variance.
In summary, we contribute a novel uncertainty quantifier that exhibits better performance in selective prediction across prompts of varying open-endedness, addressing a key limitation of the state-of-the-art in this setting.\footnote{We release all code, prompts, generations and human annotated data: \url{https://github.com/evgeniael/probar}.}

\section{Background}

Without loss of generality, we regard an LM as a probabilistic mechanism to generate responses  given a prompt. 
This mechanism is prescribed by two components: a fully-trained neural network (typically, given a prompt and an incomplete response, this NN predicts a probability distribution over candidates for the next token in the response) 
and a `sampling' algorithm (\ie, a procedure to iteratively draw tokens from next-token distributions and form a complete response; common options include ancestral or unbiased sampling \citep{bishop2006pattern}, top-k and top-p sampling \citep{fan-etal-2018-hierarchical,Holtzman2020The}, \ia). 
Given any one prompt $x$, these two components induce a conditional distribution over responses, with the the relative frequency of a response $y$ (that is, in a pool of responses obtained by repeated sampling) corresponding, in the limit, to the probability mass $p(y|x)$ the LM assigns to $y$ given $x$.\footnote{If the sampling algorithm is unbiased, this probability is in fact known in closed-form---it is the product of token-level probabilities as assigned by the underlying NN. For the distributions induced by other choices of samplers, this probability is typically available only approximately, via simulation.}
When we need to choose a single response to stand as `the LM's output' (\ie, its prediction), given a prompt, we typically introduce a decoding algorithm. A common   strategy, which we employ in this paper, is `greedy decoding' (where we form the predicted response by iteratively selecting the most probable next token), other strategies include beam search \citep{beamsearch-graves}, biased sampling algorithms (\eg, top-k, top-p, \ia) and Bayes risk decoders \citep{eikema-aziz-2022-sampling,bertsch-etal-2023-mbr}.

\begin{figure*}[h!]
    \begin{center}
        \includegraphics[width=16 cm]{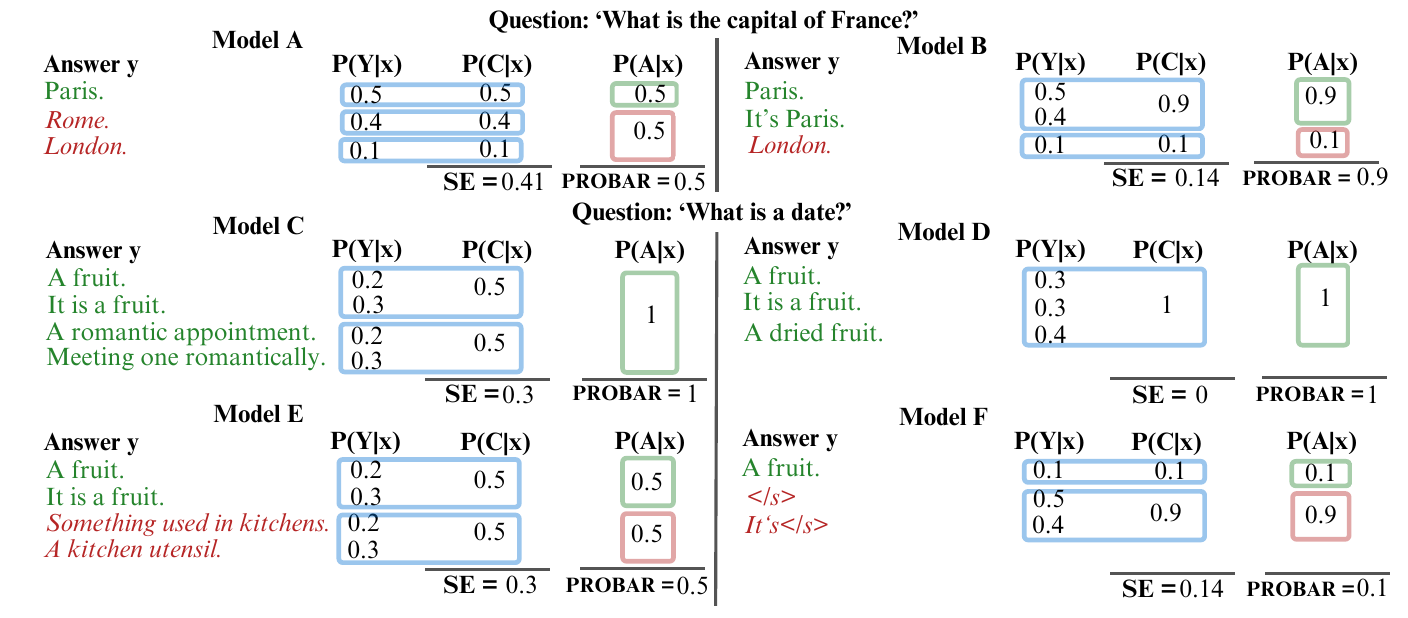}
        \caption{
        For each question, we show adequate responses (wrt a plausible interpretation of the prompt) in green  and inadequate in red (also \textit{italicised}), bounding boxes  highlight semantic clusters (relevant for SE) and adequacy/inadequacy (relevant for \probar).  For the first prompt, SE makes reasonable predictions (high/low uncertainty for model A/B, resp.) because the question strongly constrains plausible answers for their semantic content. In the second prompt, the ambiguity inherent to `date' allows for plausible answers that convey different meanings. SE makes some poor predictions (\eg, that model C and E are maximally uncertain, obscuring the fact that all answers from C are adequate; that model F is fairly certain, obscuring the fact that the dominant cluster is made of nonsensical responses), while  
        \probar makes a reasonable prediction in each case.
        }  
        \label{fig:sem_ent}
    \end{center}
\end{figure*}

\label{sec:background_measure_unc}

If two responses $y_1$ and $y_2$ are assigned probabilities such that $p(y_1|x) < p(y_2|x)$, we say the LM is less certain about $y_1$ than it is about $y_2$, given $x$.
It is often useful to express the LM's state of uncertainty wrt a prompt $x$, independently of any one response specifically. 
One such \emph{uncertainty quantifier} is Shannon entropy:
\begin{equation}
H(Y|X = x) = -\sum_{y \in \mathcal Y}p(y|x) \log p(y|x)~,
\end{equation}
where the summation is over the entire (countably infinite) space $\mathcal Y$ of possible responses. For LMs, this quantity can at best be estimated (\eg, via Monte Carlo; MC).
Entropy has been used to detect when the LM's prediction is likely wrong \cite[\emph{i.a.}][]{van-der-poel-etal-2022-mutual, rawte-etal-2023-troubling, manakul-etal-2023-selfcheckgpt,duan-etal-2024-shifting}, assuming that higher \emph{surface-form} variation amongst sampled responses is indicative of the model's inability to process the prompt and, hence, its propensity to make poor predictions. 
\citet{kuhn2022semantic} find this suboptimal, as different surface forms might still convey the same meaning. 
 To compute their \emph{semantic entropy} (SE), they map sampled responses to semantic clusters  
and estimate entropy over those:  
\begin{equation}
\operatorname{SE}(Y|X=x) \approx - J^{-1} \sum_{j=1}^{J} \log p(c_j|x) ~,
\end{equation}
where $c_1, \ldots, c_J$ are clusters formed using a model for natural language inference (NLI). In particular, bi-directional entailment between any two sampled responses (each concatenated with the prompt) signals the two responses' semantic equivalence. 
For any one cluster $c_j$, \citeauthor{kuhn2022semantic} estimate $p(c_j|x)$ by summing the closed-form probabilities of the unique responses in the cluster, despite the fact that one can only work with a (small) subset of $\mathcal Y$.
\citet{aichberger2024semantically} 
propose an improved estimator: the entropy of  the empirical distribution over clusters derived from sampled responses (that is,  $-\sum_{c_j} p(c_j|x)\log p(c_j|x)$ with $p(c_j|x)$ equal the number of responses, incl. repetitions, in cluster $c_j$ divided by sample size)---we use their SE estimator in this work. 
In essence, 
 the premise on which SE is designed and evaluated is that semantic variation signals propensity for error. 

\section{Approach}

In open-ended settings, human responses to a prompt may be semantically distinct and yet arguably valid---that can be due to prompt ambiguity or under-specification, due to annotators' varying perspectives and beliefs, amongst many other factors \citep[\eg][see section 3]{baan2023uncertainty}. Hence, in this work, we do not regard semantic variation as propensity for error. 
Figure \ref{fig:sem_ent} illustrates for hypothetical QA systems (models A--F), how semantic variation can only be argued to be indicative of error for tasks (or prompts) where the expectation is that only one answer (albeit linguistically expressible in a plethora of ways) is possible. 
For the first question ($x =$ `What is the capital of France?'), where no data uncertainty (beyond the form of paraphrasing) is expected, 
SE is successfully indicating how model B is likely more reliable than model A. However, the second ambiguous question ($x =$ `What is a date?') allows for 
multiple, semantically distinct answers (due to ambiguities inherent to `date'). 
Here, SE makes model D appear semantically certain (single cluster), while models C and E exhibit complete semantic uncertainty (uniform distribution over clusters). As it turns out, SE cannot distinguish models C and E, despite only the former having no inadequate responses.   
On the other hand, model F 
serves as an example of a pitfall of associating \emph{lack} of semantic variation with correctness: the model is semantically certain, but the dominant cluster is made of nonsensical responses. 
These examples illustrate how judging responses for their adequacy to the prompt is more informative than assessing their semantic homogeneity: it makes for an expression of confidence that has the potential to work across prompts, whether they admit more or less data uncertainty. Next, we describe our strategy to operationalise this idea.

First, we introduce a judge $a: \mathcal X \times \mathcal Y \to \{A_0, A_1\}$ that maps a prompt-response pair to a binary adequacy judgement (adequate $A_1$ or not adequate $A_0$). A judge can be a human labeller, a purposed-trained classifier or a powerful large language model \citep[`LLM as a judge';][]{li2024generation, thakur2024judging}, the latter two serving as an approximation to human labelling. 

Then, given an LM and a prompt $x$, the probability $\Pr(a(x, Y) = A_1|X=x)$ that a response is judged to be adequate $A_1$, as induced by the LM and our choice of judge, can be MC-estimated using $N$ sampled responses $y_1, \ldots, y_N$:
\begin{multline}
    \Pr(a(x, Y) = A_1|X = x)\\ \overset{\text{MC}}{\approx} \frac{1}{N}\sum_{n=1}^N [a(x, y_n)=A_1]~, 
\end{multline}
which is simply the relative frequency of adequate responses amongst the $N$ samples.\footnote{The Iverson bracket $[\text{P}]$ evaluates to $1$ if the logical predicate $\text{P}$ is True and to $0$ otherwise.} This estimate is what we refer to as \probar.

\section{Experiments}

Our experiments are designed to establish the merits of \probar as an uncertainty quantifier apt to detect which prompts the LM is likely to form correct predictions for, associating low uncertainty (or high confidence) with those, and, conversely, which prompts the LM is likely to form incorrect predictions for, associating high uncertainty (or low confidence) with those.
\paragraph{Tasks and Datasets.}

We use Abg-COQA \citep{guo2021abg}, a reading comprehension QA (RCQA) dataset; where a passage is accompanied by one or more rounds of questions. We separate the observations in this dataset in two portions, one 
where the last question (in the round) is ambiguous and, given the passage, multiple answers are plausible, and another without ambiguity. This comprises 994 context-ambiguous question pairs (741, 130, 123 in training, development and test sets, resp.) and 253 non-ambiguous prompts (130 and 123 in development and test sets, resp.). 
We also report experiments on AmbigQA \citep{min-etal-2020-ambigqa}, which is a knowledge based QA (KBQA) dataset. We focus on the 1070 questions labelled as ambiguous from the development set.\footnote{We used the version from \url{https://huggingface.co/datasets/sewon/ambig_qa/tree/main/full}, which only contains training and development sets.}
 Finally, we also experiment with next-word prediction (NWP), as a task that portrays high plausible variability.  As prompts for this task, we use 100 randomly chosen contexts from Provo Corpus \citep{luke2018provo}, a dataset of passage prefixes in English from various sources. Each prefix from the dataset contains, on average, 40 human-labelled next word continuations; which provide us with plausible reference answers for the NWP task.

\paragraph{Models.}
We 
generate responses from OPT models \cite{zhang2022opt} of varying sizes (2.7B, 6.7B, 13B and 30B). For each prompt (and model size), we obtain the greedy decoding (which stands as the model prediction for the prompt) and we also obtain 10 unbiased samples (for uncertainty quantification). 
For Abg-COQA, the prompt is comprised of a passage, previous rounds of QA pairs and the ambiguous question. 
For AmbigQA, the prompt is made of 10-shot question-answer pairs as examples, similar to \citet{kuhn2022semantic}, followed by the ambiguous question. 
For both QA datasets, a sampled response is obtained by sampling tokens iteratively until an end-of-sequence token is sampled (or until a maximum length of 150 tokens).
For Provo Corpus, the prompt is made of a passage prefix. We follow \citet{ilia-aziz-2024-predict} and sample a response (\ie, a choice for the next word) by sampling subword tokens iteratively until a complete word is detected. To simulate cases in which the model would not be able to respond to the prompt,\footnote{We need to simulate those cases since we can expect autoregressive LMs pre-trained on English corpora to perform the NWP task well intrinsically, due to their training objective.} we randomly replace each context by a same-length context.  The model prompts for all tasks are detailed in \Cref{appendix:model_prompts}.

\paragraph{Metrics.} We evaluate all uncertainty quantifiers for their usefulness in selective prediction \citep{kamath-etal-2020-selective, varshney-etal-2022-investigating}, where, for each prompt, the model's prediction is accepted or rejected as a function of the uncertainty quantifier (\eg, reject responses generated under more uncertainty than the user tolerates). 
For the QA tasks, we regard the greedy decoding as the LM's prediction, as typically done for QA \citep{kuhn2022semantic}.
For NWP, we choose (at random) one of the sampled responses and regard that as the LM's prediction. 
There are different metrics that summarise precision vs. coverage tradeoffs at varying uncertainty thresholds. 
In Section \ref{sec:results}, we report area under the receiver operating characteristic curve (AUROC).\footnote{Interpretable as the rate at which a randomly chosen correct prediction is made under lower uncertainty (or higher confidence) than a randomly chosen incorrect one.} Higher AUROC is desirable, with the ideal uncertainty quantifier achieving an AUROC score of 1. 
Additionally, in \Cref{appendix:risk_cov_plots}, we visualise selective precision (fraction of correct over predicted instances) versus coverage (fractions of predicted over all instances) plots.
These evaluation metrics require a notion of correctness by which to criticise the LM's prediction. In all of our datasets, we have a reference set of plausible answers to which we can compare the LM prediction. For the QA datasets, 
similar to \citet{lin2024generating}, we let \texttt{gpt3.5-turbo} (via the OpenAI API) determine whether the LM prediction is correct: we prompt it to generate true if the prediction is plausible given the question, the passage (if relevant) and the reference answers. 
For Provo Corpus, we regard the LM prediction as correct if it exactly matches one of the prefix's reference answers. 

\paragraph{Baselines.}
As baselines for comparison, we employ entropy (E), estimated via MC, semantic entropy (SE) estimated following \citet{aichberger2024semantically}, and a variant of P(True) \citep{kadavath2022language}. 
SE requires a clustering algorithm; we follow the strategy by \citet{kuhn2022semantic}, using bidirectional entailment. Specifically, for Abg-COQA and AmbigQA, we adopt Deberta-Large \cite{he2020deberta} as the NLI model. For Provo, instead, we employ an LM, \texttt{Mistral-Nemo-Instruct-2407} \citep{AI_2024}, to act as an NLI classifier,\footnote{We opt for an LM as NLI models are trained on complete sentences rather than incomplete ones (like passage prefixes).} prompted similarly to the implementation by \citet{farquhar2024detecting}---details in \Cref{appendix:sem_equiv_model}.
When computing AUROC, we noticed for both E and SE that the scale of entropy values varies across prompts, as the number of elements in the distributions' support varies per instance (\ie, the number $M$ of unique responses for E, and the number $J$ of clusters for SE), which leads to some misleading AUROC values. We adjust for that by normalising entropy (E or SE) by its theoretical upperbound in each case, 
and transforming it to a confidence score: $1-\frac{\operatorname{H}(Y|X=x)}{\log M}$ for E and $1-\frac{\operatorname{SE}(Y|X=x)}{\log J}$ for SE; these are denoted Norm.E and Norm.SE in plots.  
We adapt P(True) to prompt for adequacy instead of correctness (details in \Cref{appendix:p_adequate}); this variant is denoted P(Adequate) in plots.

\begin{figure*}[!h]
\centering
\begin{subfigure}[b]{\textwidth}
    \centering
    \includegraphics[width=11cm]{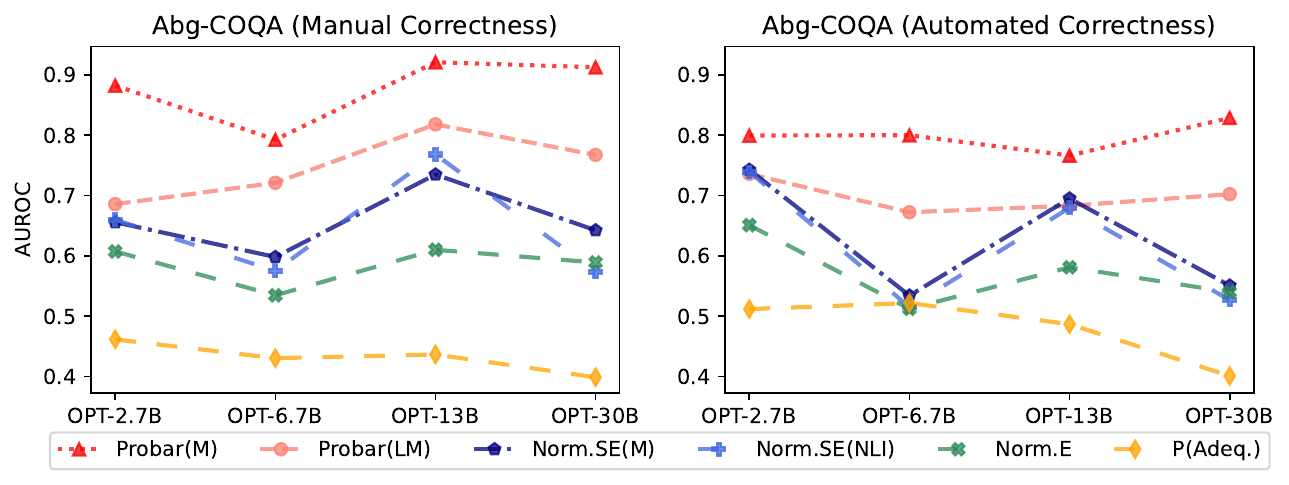}
    \caption{AUROC values for AbgCOQA's 50 manually annotated contexts. On the left, AUROC is computed using manual correctness annotations for the greedy response; on the right, correctness of the greedy was automated using \texttt{gpt3.5-turbo}.}
    \label{figure:abgcoqa_auroc_manual}
\end{subfigure}
\centering
\begin{subfigure}[b]{\textwidth}
    \includegraphics[width=16cm]{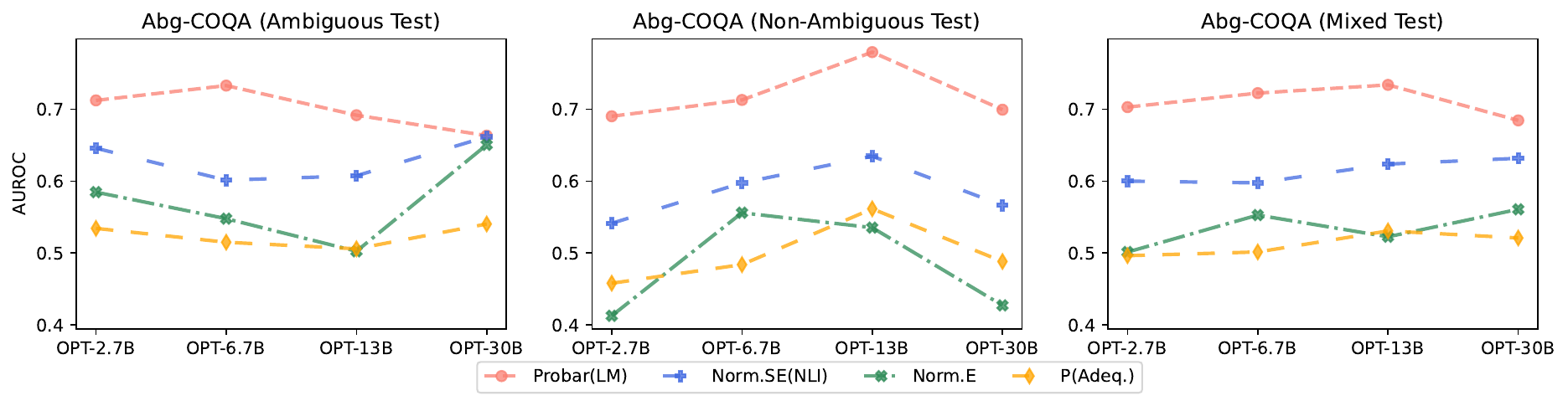}
    \caption{AUROC values for AbgCOQA's test set for ambiguous prompts (left), non-ambiguous prompts (middle) and their aggregated set (right); 123, 123 and 246 prompts respectively. Correctness of the greedy was automated using \texttt{gpt3.5-turbo}.}
    \label{fig:abgcoqa_auroc_test}
\end{subfigure}

\caption{AUROC values for various quantifiers for Abg-COQA.}
\label{fig:auroc_vs_model}
\end{figure*}

\paragraph{\probar implementation.}
For each task, \probar requires an adequacy classifier. 
For Abg-COQA, we considered two different strategies: (1) approximate adequacy judgements using an NLI model, where the passage is regarded as the premise and an affirmative sentence comprised from the question-response pair is regarded as the hypothesis,  
and (2) have an LM perform adequacy judgements, with the passage and a question-answer pair provided as context and the LM prompted to generate true if the answer to the question is plausible given the passage, or false if not.
For AmbigQA, we prompt an LM to generate true if a response is adequate given the question with respect to the LM's own training data / parametric knowledge.
For NWP, we prompt an LM to generate true if a response is plausible given the context and false otherwise. 
No adequacy classifier internal to \probar has access to reference answers.
\footnote{Details about models and prompts in \Cref{appendix:adequacy_classifiers}.}

\paragraph{Manual Evaluation.} 
\label{subsec:eval_algorithms}
We perform human evaluation of a number of automated components. In particular, on Abg-COQA, we evaluate clustering algorithms for SE, adequacy classifiers for \probar, as well as the LLM component of the evaluation protocol (which assesses the correctness of a selected answer in relation to the available reference answers). 
We randomly sample 50 context-ambiguous question pairs from the training set of Abg-COQA and manually annotate all responses sampled from OPT-models for their semantic equivalence and adequacy. 
We use these to characterise upperbounds on the performance we can expect from SE and \probar in this setting.
We also used the manual adequacy labels to optimise design choices for our classifiers on F1---the best performing classifier (which we use across the main experiments) is based on \texttt{Mistral-Nemo-Instruct-2407} \citep{AI_2024} (see \Cref{appendix:adequacy_classifiers}, prompt \texttt{LM-1-Step-Plausible}, for all details).
For the same 50 context-ambiguous question pairs, we also manually annotate OPT models' greedy outputs for their correctness. This allows us to evaluate \texttt{gpt3.5-turbo}'s performance in automating correctness decisions when computing AUROC. 
These labels also allow us to assess the uncertainty quantifiers' AUROC performance when no errors in the automated components of the evaluation setup occur (measuring \emph{actual} performance of the implemented quantifiers).
Last, we labelled generations for 10 contexts from the Provo Corpus in order to assess the performance of \probar's adequacy classifier in NWP.
Details on annotations and evaluation results in \Cref{appendix:manual_annotations} and \ref{appendix:eval_adeq_classifiers}.

\begin{figure*}
\centering
\includegraphics[width=11cm]{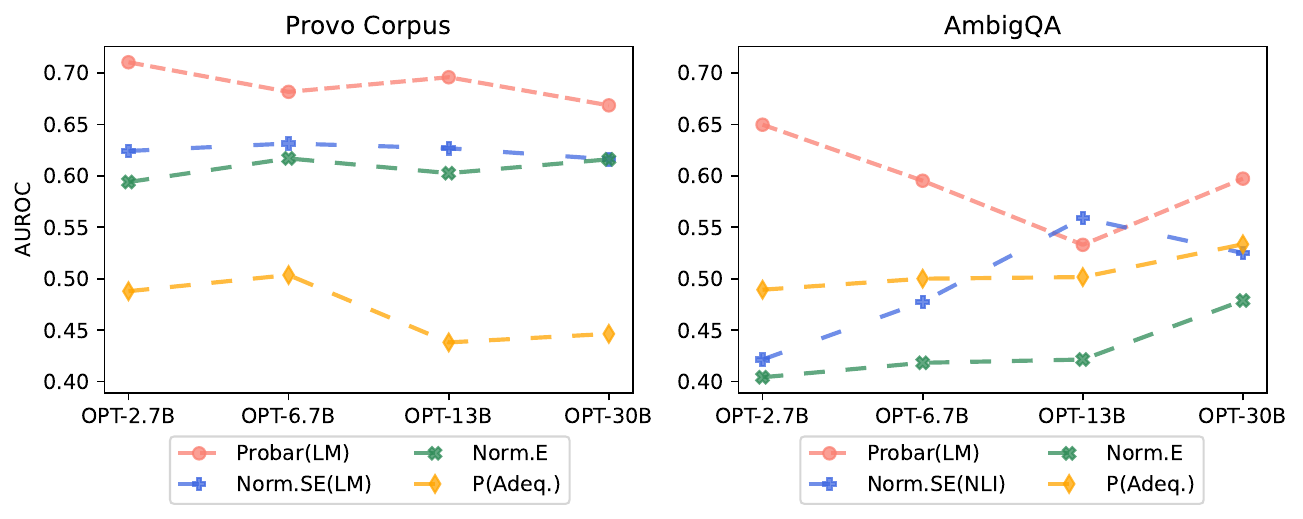}
\hfill
\caption{AUROCs for Provo Corpus (left) and AmbigQA (right) tasks. For Provo Corpus (200 prompts), correctness was assessed via exact matching of a sampled response to human references (AUROC is as an average over 5 runs). For AmbigQA (1070 prompts), correctness of the greedy was assessed by \texttt{gpt3.5-turbo}.}
\label{fig:provo_auroc_vs_model}
\end{figure*}

\section{Results}
\label{sec:results}

\paragraph{Quantifiers' upperbounds on Abg-COQA.}
In \Cref{figure:abgcoqa_auroc_manual}, we concentrate on the subset of AbcCOQA for which we obtained manual labels (for semantic clusters in SE, adequacy judgements in \probar, and the correctness of greedy responses in the evaluation protocol). In both plots, we observe AUROC values for different uncertainty quantifiers across OPT model sizes, with SE's and \probar's internal decisions replaced by human judgement.  
On the left, AUROCs are computed using hand-labelled correctness for the greedy responses, hence we observe results in a setting free of errors in the evaluation protocol. 
\probar(M) and Norm.SE(M) represent the `upper-bounds' for SE and \probar, respectively, computed using the manual adequacy and semantic equivalence annotations. \probar outperforms all baselines, both its upperbound \probar(M) and its practical implementation \probar(LM). Beyond the fact that \probar(M) surpassed Norm.SE(M) for all OPT models by a large margin, it is noteworthy that \probar(LM) surpassed Norm.SE(M).\footnote{In \Cref{appendix:sample_for_auroc} we study other choices of decoders for the selected model-generated response used in the AUROC computation,  with similar patterns observed.} 
On the right plot of \Cref{figure:abgcoqa_auroc_manual}, we see the results for the same subset, but using \texttt{gpt3.5-turbo} to automate the evaluation protocol (\ie, assess correctness of the greedy responses). 
We can see that errors in evaluation have a greater impact on more informative quantifiers (\emph{i.e.} with higher AUROC values). Those get negatively impacted by misclassifications of some of the plausible decodings they accept with high confidence, while worse uncertainty quantifiers, which wrongly abstain from answering, dodge these evaluation errors. 

\paragraph{Quantifiers' practical performance on AbgCOQA.}
We now turn to the test set, with all relevant components automated (semantic clusters in SE, adequacy judgements in \probar, and the correctness of greedy responses in the evaluation protocol).
Across all settings in \Cref{fig:abgcoqa_auroc_test} (ambiguous prompts, non-ambiguous prompts and their combination), \probar outperforms the baselines. In one case, \probar, Norm.SE and Norm.E all achieve a similar AUROC score. This can occur when OPT models were rather confident in a single response (which was ultimately also correct); circumstances under which Norm.E and Norm.SE could predict model correctness well. 
 In the non-ambiguous setting, we believe \probar's superior performance can be attributed to its ability to detect `confident' errors (where a model is confident in a response, which is ultimately incorrect). To analyse our results' variance, we perform a bootstrap analysis in \Cref{appendix:ensemble_analysis}, where subsets of prompts are repeatedly sampled to compute AUROCs. Moreover, we show the precision-coverage plots in \Cref{appendix:risk_cov_plots}, that generally paint a similar picture (\probar outperforms other quantifiers).

\paragraph{AmbigQA.}
We observe the results for the ambiguous prompts from AmbigQA (\Cref{fig:provo_auroc_vs_model}, right), where \probar outperforms other baselines, except for OPT13B. We analyse how well OPT-models model plausible variability to assess whether their similar performance is due to the assessed OPT models inability to capture all plausible answers in their responses. In Appendix \Cref{fig:e_se_ambigqa}, we observe the variation among is largely due to paraphrases (as the histograms of E and SE reveal). This is in contrast to variation observed in AbgCOQA and Provo Corpus, where variation among semantically distinct responses is more apparent  (Appendix \Cref{fig:e_se_abgcoqa} and \ref{fig:e_se_provo}). 

\paragraph{Provo Corpus.}
In \Cref{fig:provo_auroc_vs_model} (left), \probar outperforms all other quantifiers, robustly informing us when the model can respond to a prompt, despite of the heightened data uncertainty in the NWP task.

\section{Discussion}

We demonstrated how semantic variation need not robustly predict propensity for error.
This carries a broader implication: a distribution over responses given a prompt and its basic properties, such as its spread or shape, need not be informative of the correctness of a specific model-generated response.
This does not go to say that SE or other metrics based on the same premise are flawed, but their effectiveness is arguably limited to settings (tasks or prompts) where we strongly expect correct responses to differ at most in \emph{how} they convey a single meaning (\ie, paraphrastic variation) and not in \emph{what} meaning they convey---which excludes settings of varying (or unknown) open-endedness.
When considering uncertainty metrics to employ, one needs to carefully consider the settings of their problem, task and data. 
In more open-ended tasks with high data uncertainty, \emph{e.g.} dialogue, open-ended QA, story generation etc., 
they could consider employing a metric like \probar. 
We demonstrate how \probar  better informs us whether the model can reliably respond to a prompt, regardless of the prompt's open-endedness. 
We also envision potential for finding complementary information in different uncertainty quantifiers. For example, simultaneously high SE and high \probar might detect prompts for which the LM models plausible variability well, while simultaneously high SE and low \probar might detect prompts about which the model is rather ignorant. Future work can investigate such relationships. 
Moreover, \probar could power confidence-aware decoding. 
In \Cref{appendix:probar_decoding} we take some initial steps in this direction (by regarding one of the responses that was deemed adequate as the chosen LM prediction) to seed future research.

\section{Related Work}
Various methods were proposed to quantify aspects of uncertainty: some based on calibration \citep{kumar2019calibration, jiang-etal-2021-know}, some on LMs' predicted probabilities \citep{varshney2023stitch, bakman-etal-2024-mars, lin-etal-2024-contextualized}, and some on the LM's own uncertainty verbalisation \citep{kadavath2022language, mielke-etal-2022-reducing}. 

\paragraph{Consistency based Uncertainty Estimation.}
Another method to assess a model's uncertainty is based 
on summarising distributional information from LMs along different `consistency' dimensions. 
Beyond entropy \citep{fomicheva-etal-2020-unsupervised, malininuncertainty, xu-etal-2020-understanding-neural, xiao-wang-2021-hallucination, rawte-etal-2023-troubling} and semantic entropy \citep{kuhn2022semantic, farquhar2024detecting}, other work computes semantic uncertainty with diffeent methods:
\citet{aichberger2024semantically} use importance sampling; \citet{cheng-vlachos-2024-measuring} use similarity-sensitive entropy; \citet{nikitin2024kernel} use semidefinite unit trace kernels to model semantic similarities;  \citet{chen2024inside} use the eigen-values of embeddings of sampled responses and 
\citet{rabinovich-etal-2023-predicting} use the pairwise similarity of embeddings of responses.
\citet{scherrer2024evaluating} map sampled responses to semantic actions 
and \citet{aina-linzen-2021-language} map to responses' interpretations before computing uncertainty metrics.
Rather than assessing consistency among responses, \citet{manakul-etal-2023-selfcheckgpt} assess consistency of responses to a specific response, while  
\citet{chen-mueller-2024-quantifying} and \citet{xiong2023can} compute confidence scores using sampling of models' responses and their confidence scores.
Alternatively, rather than repeatedly sampling, responses' 
some work perturbs inputs. 
For instance, \citet{zhao-etal-2024-knowing} analyse divergence of responses on rephrased inputs to detect hallucinations, while \citet{tonolini-etal-2024-bayesian} compute probabilities with a weighted ensemble of rephrased task instructions.
Similarly, others compute consistency-based metrics based on responses from paraphrased prompts \citep{elazar-etal-2021-measuring, fierro-sogaard-2022-factual, raj2023semantic, yang2024just,gao-etal-2024-spuq, li2024benchmarking}. 
Beyond sampling outputs from one model, some assess cross-model samples \citep{zhang-etal-2023-sac3, zhang-etal-2024-luq}. 
Other dimensions of consistency 
investigated include logical \citep{jang-etal-2022-becel}, concept \citep{sahu2022unpacking} and reasoning path consistency \citep{wang2023self}.
\probar joins this stream of work; assessing how consistently an LM generates samples deemed as adequate by a task-specific classifier.

\paragraph{Decomposing Uncertainty} Some work attempts to detect when a model will err by decomposing total uncertainty into aleatoric and epistemic. 
\citet{gao-etal-2024-spuq} quantify those aspects by sampling responses for pertrubed inputs; \citet{hou2023decomposing} by ensembling responses with input clarifications and \citet{ling-etal-2024-uncertainty} by sampling responses under different in-context demonstrations.
\citet{cole2023selectively} prompt LMs repetitively to generate interpretations of a question and responses and aggregates into confidence scores, decomposing denotational and epistemic uncertainty. 
\citet{kuhn2022clam} and \citet{zhang2023clarify} detect ambiguous questions to generate clarification questions; targeting uncertainty caused by ambiguity. 
\citet{yadkori2024believe} compute a lower bound on epistemic uncertainty using iterative sampling. 
Rather than decomposing uncertainty or detecting when an input has high aleatoric uncertainty, \probar aims to indicate when a model can reliably respond to a prompt, regardless of potential data uncertainty in the input.

\section{Conclusion}
We demonstrate how semantic variation is not a robust signal for assessing a model's ability to respond to a prompt in settings where data uncertainty, beyond the form of paraphrasing, is present; particularly relevant to open-ended tasks. 
We propose \probar, an uncertainty quantifier based on the probability the LM generator assigns to the set of responses that are adequate to the prompt; we approximate adequacy judgements using a classifier and use an empirical estimate of the adequacy probability. 
We demonstrate the efficacy of \probar in selective prediction with OPT models in two QA datasets as well as next-word prediction, all exhibiting prompts of varying open-endedness.

\section*{Limitations}

Regarding limitations of our study, we have identified the following:
\probar, as most sampling-based uncertainty quantification methods are generally computationally expensive. However, we believe that the benefits of obtaining useful and reliable uncertainty signals outweigh the computational costs. Nevertheless, we conduct an ablation study in \Cref{appendix:num_samples_ablation}, where we find that 5 samples would generally be sufficient to obtain results similar to our experiments with 10 samples. Beyond that, future research could investigate whether one could learn to predict \probar (similar to how \citet{kossen2024semantic} predict SE).
Furthermore, \probar requires a task-specific classifier. Essentially, if one wishes to have an adequacy classifier for tasks different than the ones examined, they would need to construct and evaluate their own (possibly inspired by how we did it).
Obviously \probar is also highly dependent on the quality of the adequacy classifier---the better the classifier's performance, the more informative the quantifier will be. This sets as vital that one thoroughly evaluates the adequacy classifier in their setting of interest, given relevant data. That being said, due to limited resources, we only could manually annotate responses to perform a thorough evaluation of the adequacy classifiers from the RCQA task (and a smaller scale evaluation for the NWP task). The classifiers for the KBQA task was not explicitly evaluated for their performance, and we only `implicitly' validate their performance through the improved AUROC performance of \probar.
At the same time, we only evaluated \probar in a setting where only short responses were expected. If one needed to employ an adequacy classifier in a QA setting with longer generations expected, they would need to re-assess the classifier's performance.
At the same time, within our study, adequacy of the response was assessed as a binary decision reflecting whether the response is a plausible answer to the question given a passage (RCQA), the classifier's training data (KBQA) or a plausible continuation to a context (NWP task). However, within different applications and domains, the notion of adequacy can be as general or as fine-grained as one defines it (\eg, one might consider a response as adequate only if it contains a plausible answer, is grammatically coherent and does not contain toxic text). In this case, they would need to align the training and evaluation of the corresponding adequacy classifier according to their needs.

\section*{Acknowledgements}

\begin{wrapfigure}{l}{0.12\linewidth}
\vspace{-13pt}
\includegraphics[width=0.08\textwidth]{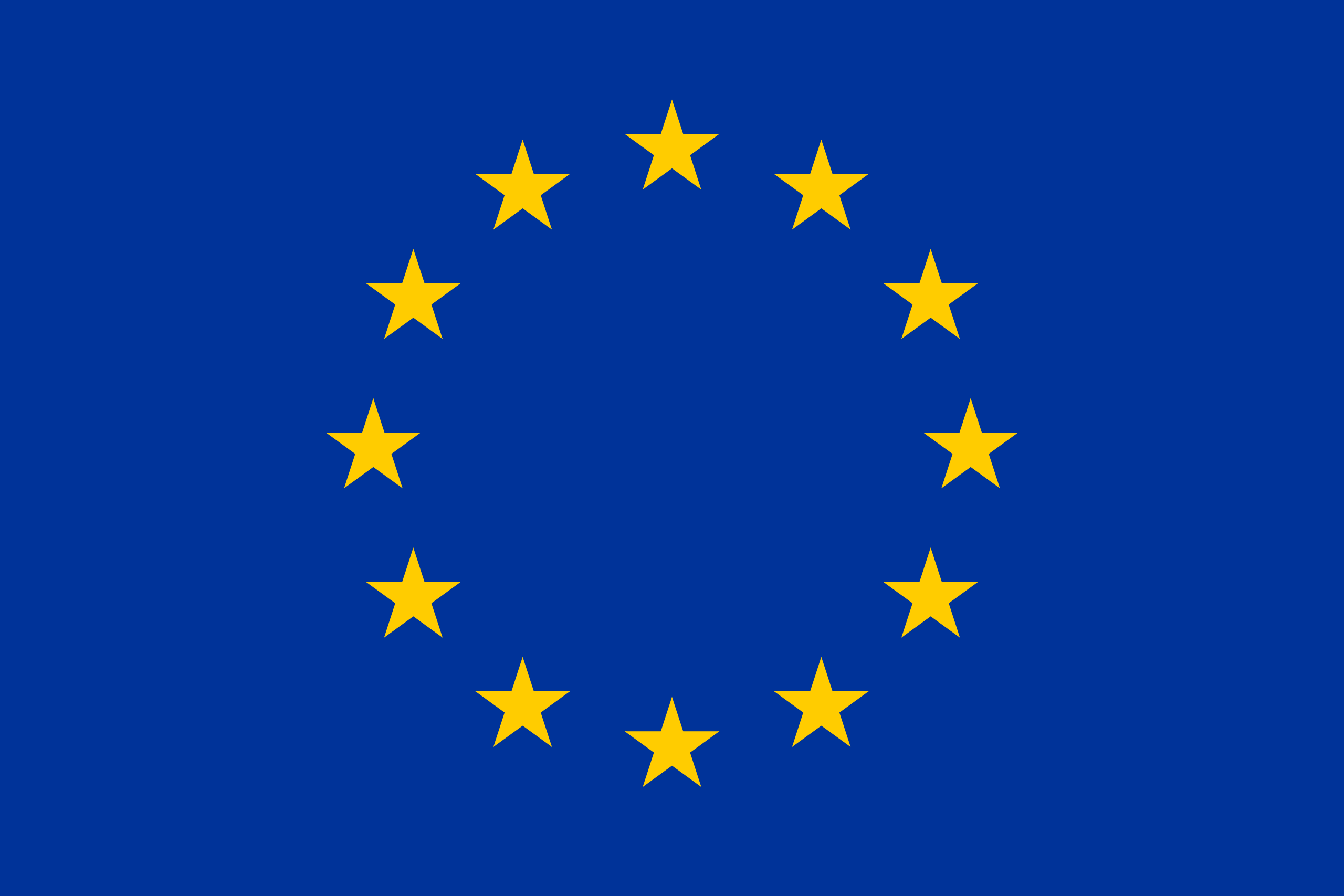}
\end{wrapfigure}

This project has received funding by the EU’s Horizon Europe research and innovation programme (grant agreement No. 101070631, UTTER).

\bibliography{anthology,custom}

\appendix

\section{Experimental details}

\subsection{Analysed models' Prompts}
\label{appendix:model_prompts}
For AmbigQA, we use a 10-shot prompt, with 10 example question-answer pairs from the datasets' training set:

\texttt{Question: When did the simpsons first air on television? Answer: April 19, 1987 Question: Who played george washington in the john adams series? Answer: David Morse Question: What is the legal age of marriage in usa? Answer: 18 years of age Question: Who starred in barefoot in the park on broadway? Answer: Elizabeth Ashley Question: When did the manhattan project began and end? Answer: Began 1939, end 1946 Question: When did the frozen ride open at epcot? Answer: June 21, 2016 Question: Name the landforms that form the boundaries of the peninsular plateau? Answer: Aravali Range, Satpura Range, Vindhyan Range Question: When was the last time uga won a national championship? Answer: 1980 Question: Who sing play that funky music white boy? Answer: Rob Parissi Question: When was the first airplane used in war? Answer: Bl\u00e9riot XI Question: <AMBIGUOUS\_QUESTION> Answer:}

\begin{figure}[t]
    \begin{center}
        \includegraphics[width=7 cm]{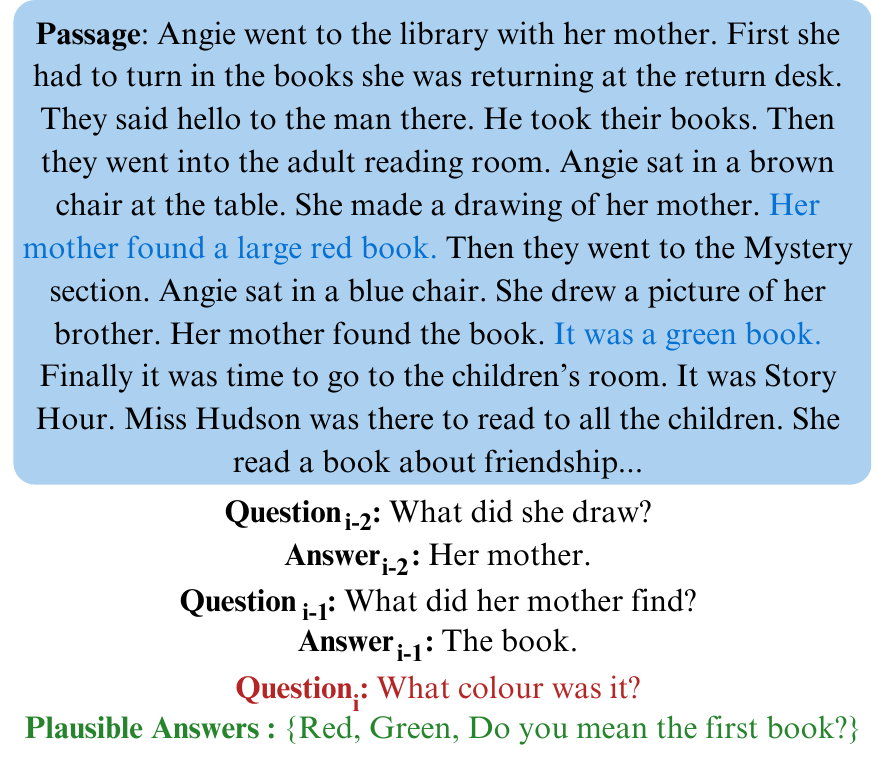}
        \caption{Example of an ambiguous instance from Abg-COQA.}  
        \label{fig:abg_coqa_example}
    \end{center}
\end{figure}

For Abgcoqa, each datapoint is comprised by a passage, some previous rounds of question-answer pairs and lastly, the ambiguous question, as in the example of \Cref{fig:abg_coqa_example}. Hence, the prompt for each question is constructed as follows:

\texttt{Context: <PASSAGE>}

\texttt{Question: <PREVIOUS\_QUESTION\_1>}

\texttt{Answer:<PREVIOUS\_ANSWER\_1>}

\texttt{Question: <PREVIOUS\_QUESTION\_2>}

\texttt{Answer:<PREVIOUS\_ANSWER\_2>}

\texttt{Question:<AMBIGUOUS\_QUESTION>}

\texttt{Answer:}

As these models are pre-trained and tend to continually generate question-answer pairs following a few-shot example, we add to the end-of-sequence tokens, beyond the `$<EOS>$' token, the tokens corresponding to the words `question', `answer', `Question', `Answer', `question:', `answer:' and `.'.

To obtain generations we sampled unbiasedely. From Hugging Face, the relevant parameters are:

\texttt{do\_sample=True}

\texttt{num\_beams=1}

\texttt{temperature=1}

\texttt{top\_p=1}

For the greedy decoding, the parameters are:

\texttt{do\_sample=False}

\texttt{num\_beams=1}

\subsection{Semantic equivalence assessment}
\label{appendix:sem_equiv_model}
For computing SE, we need to operationalise an NLI based algorithm, as explained in Section 2.2. For QA experiments, Deberta-Large was used. However, for NWP, the inputs (prefix contexts) were incomplete sentences – which is unlike the training of NLI classifiers such as Deberta-Large. Hence, we prompt a LM, in this case \texttt{Mistral-Nemo-Instruct-2407}
to act as an NLI classifier. The prompt used for assessing bi-directional entailment is:

\texttt{You are presented with two strings, String 1 and String 2. Generate True if String 1 semantically entails String 2 and False otherwise. Only generate True or False. String 1:'<STRING1>' String 2:'<STRING2>'.}

\subsection{Adequacy assessment for P(Adeq.)}
\label{appendix:p_adequate}
We adjust \citeauthor{kadavath2022language}'s P(True) to a, relevant in our case, P(Adequacy). For that, we prompt the original analysed models (OPT2.7B-30B) as follows:

\begin{itemize}
    \item \textbf{Prompt for RCQA task:}

    \texttt{Context: '<PASSAGE>'}

    \texttt{Question: '<QUESTION>'}
    
    \texttt{Here are some brainstormed ideas: '<SAMPLED\_RESPONSES>'}
    
    \texttt{Possible answer: <ANSWER>}
    
    \texttt{Is the possible answer:}
    
    \texttt{(A) Plausible}
    
    \texttt{(B) Not Plausible}
    
    \texttt{The possible answer is:}

    \item \textbf{Prompt for KBQA task:}
    
     \texttt{Question: '<QUESTION>'}
    
    \texttt{Here are some brainstormed ideas: '<SAMPLED\_RESPONSES>'}
    
    \texttt{Possible answer: <ANSWER>}
    
    \texttt{Is the possible answer:}
    
    \texttt{(A) Plausible}
    
    \texttt{(B) Not Plausible}
    
    \texttt{The possible answer is:}

    \item \textbf{Prompt for NWP task:}

    \texttt{Context: '<CONTEXT>'} 
    
    \texttt{Here are some brainstormed continuations: <CONTINUATIONS>}
    
    \texttt{Possible continuation: <GREEDY>} 
    
    \texttt{Is the possible continuation:}
    
    \texttt{(A) Plausible}
    
    \texttt{(B) Not Plausible}
    
    \texttt{The possible continuation is:}

\end{itemize}

We assess the log-probabilities assigned by the models to the sequences \texttt{P(` (A)'|prompt)} and \texttt{P(` (B)'|prompt)}, using the model's output layer's logits, and we normalise to obtain the probability assigned to \texttt{P(` (A)')}, which is used as a proxy to \texttt{P(Adequate)}.

\subsection{Adequacy assessment for \probar}
\label{appendix:adequacy_classifiers}
The function mapping a response to an adequacy assessment is task-specific. Hence, for the different tasks we assessed different prompting techniques and different models.

\begin{itemize}
    \item \textbf{Next word prediction}

    \begin{enumerate}
        \item \textbf{Method}: LM as classifier
    
    \textbf{Model}:  \texttt{Mistral-Nemo-Instruct-2407} (referred to as \texttt{Mis12B})

    \textbf{Prompt}: \texttt{You are presented with a piece of text and a continuation. Generate True if the continuation is a plausible continuation to the context and False if it is not a plausible continuation. By plausible, I mean that when concatenating the continuation to the text, the text will remain grammatically correct and comprehensible. Text:'<CONTEXT>' Continuation:'<WORD>' Answer:}
    \end{enumerate}
    \item \textbf{Knowledge based Question Answering}

    \begin{enumerate}
        \item \textbf{Method}: LM as classifier
    
    \textbf{Model}: \texttt{Mistral-Small-Instruct-2409} (referred to as \texttt{Mis22B})
    
    \textbf{Prompt:} \texttt{You are presented with a question and an answer. Generate True if the answer is a plausible response to the question with respect to your training data and False if not. Only generate True or False. Question:'<QUESTION>' Answer:'<ANSWER>'.}
    \end{enumerate}
    
    \item \textbf{Reading Comprehension Question Answering}

    \begin{enumerate}
        \item \textbf{Method}: NLI-based classifier
        
        A NLI model is given as a premise the passage and as a hypothesis an affirmative sentence comprised from the question-response pair, generated using a LM. If the entailment probability is higher than the sum of probabilities of neutrality and contradiction, then the response is considered as adequate. 
        
        \textbf{Model}: \texttt{Deberta-Large} \citep{he2020deberta} and \texttt{Roberta-large}, fine-tuned on a more challenging dataset collected via an iterative, adversarial human-and-model-in-the-loop procedure \citep{nie-etal-2020-adversarial}. The former method is referred to as \texttt{NLI-Easy} and the latter as \texttt{NLI-Hard}.
        \item \textbf{Method}: NLI-based classifier (using a LM for the NLI task)
        
        \textbf{Model}: \texttt{Mistral-7B-Instruct-v0.2} and \texttt{Mistral-Nemo-Instruct-2407}  (referred to as \texttt{Mis7B} and \texttt{Mis12B} respectively).

        \textbf{Prompts}:
        \begin{enumerate} 
            \item \emph{NLI-LM-2}: \texttt{You are given a premise and a hypothesis. Generate True if the hypothesis is entailed by the premise and False if the hypothesis is not entailed by the premise. Only generate True or False. Premise: '<PASSAGE>' Hypothesis:'<SENTENCE>'}
            \item \emph{NLI-LM-3}: \texttt{You are given a premise and a hypothesis. Generate Entailment if the hypothesis is entailed by the premise, Contradiction if the hypothesis is contradicted by the premise and Neutral if not possible to determine. Only generate Entailment, Contradiction or Neutral. Premise: '<PASSAGE>' Hypothesis:'<SENTENCE>'}
        \end{enumerate} 
        \item \textbf{Method}: LM as adequacy classifier

    We prompt a LM to act as a classifier, given a document (passage) and a claim (affirmative sentence created from a question response pair from a LM), or the question-answer pair directly.

    \textbf{Models}: \texttt{Mistral-7B-Instruct-v0.2}, \texttt{Mistral-Nemo-Instruct-2407} and \texttt{Mistral-Small-Instruct-2409}, referred to as \texttt{Mis7B}, \texttt{Mis12B} and \texttt{Mis22B} respectively.

    \textbf{Prompts}:
        \begin{enumerate} 
            \item \emph{LM-1-Step-Support}: \texttt{You are presented with a document, a question based on the document and an answer to the question. Generate True if the answer to the question is supported by the document and False if the answer to the question is not supported by the document. Only generate True or False. Document: '<PASSAGE>' Question:'<QUESTION>' Answer:'<ANSWER>'.}

            \item \emph{LM-1-Step-Plausible}: \texttt{You are presented with a document, a question based on the document and an answer to the question. Generate True if the answer to the question is plausible given the document and False if the answer to the question is not plausible given the document. Only generate True or False. Document: '<PASSAGE>' Question:'<QUESTION>' Answer:'<ANSWER>'.}

            \item \emph{LM-1-Step-COT}: \texttt{You are presented with a document, a question based on the document and an answer to the question. Generate True if the answer to the question is plausible given the document and False if the answer to the question is not plausible given the document. Generate your intermediate reasoning steps before generating your final answer. Document: '<PASSAGE>' Question:'<QUESTION>' Answer:'<ANSWER>' Reasoning:}

            \item \emph{LM-2-Steps-Support}: \texttt{You are presented with a document and a claim. Generate True if the claim is supported by the document and False if the claim is not supported by the document. Only generate True or False. Document: '<PASSAGE>' Claim:'<SENTENCE>'}
            
            \item \emph{LM-2-Steps-NoContradiction}: \texttt{You are presented with a document and a claim. Generate True if the claim is not contradicted by the document and False if the claim is contradicted by the document. Only generate True or False. Document: '<PASSAGE>' Claim:'<SENTENCE>'}
            
            \item \emph{LM-2-Steps-Support-Few-Shot}: \texttt{You are presented with a document and a claim. Generate True if the claim is supported by the document and False if the claim is not supported by the document. Only generate True or False, as in the following example: <EXAMPLE>}
            
            \texttt{Document: <PASSAGE> Claim: '<SENTENCE>' Prediction:}
            
        \end{enumerate}
        
        In the cases where an affirmative \texttt{<SENTENCE>} was needed, \texttt{Mis7b} was prompted to generate it given a question-answer pair: 
        
        \texttt{Turn a question-answer pair to a declarative sentence. Only output the sentence and nothing else. Question: '<QUESTION>' Answer: '<ANSWER>'}

        To obtain a decision from a model's response to the prompt, we do the following: if the lower-cased response contains the string 'true' then we consider the response as adequate; if the lower-cased response contains the string 'false' then we consider the response as non-adequate; if the lower-cased response contains neither the strings 'true' nor 'false', or both of these strings, we dismiss the response when computing \probar for the corresponding prompt.
    \end{enumerate}
\end{itemize}

\begin{table*}[]
\resizebox{16cm}{!}{
\begin{tabular}{l|rrrr|rrrr|rrrr}
 & \multicolumn{4}{c|}{\textbf{NLI-Easy}} & \multicolumn{4}{c|}{\textbf{NLI-Hard}} & \multicolumn{4}{c}{\textbf{NLI-LM-2(Mis7B)}} \\ \cline{2-13} 
 & \multicolumn{1}{l|}{\textbf{OPT2.7B}} & \multicolumn{1}{l|}{\textbf{OPT6.7B}} & \multicolumn{1}{l|}{\textbf{OPT13B}} & \multicolumn{1}{l|}{\textbf{OPT30B}} & \multicolumn{1}{l|}{\textbf{OPT2.7B}} & \multicolumn{1}{l|}{\textbf{OPT6.7B}} & \multicolumn{1}{l|}{\textbf{OPT13B}} & \multicolumn{1}{l|}{\textbf{OPT30B}} & \multicolumn{1}{l|}{\textbf{OPT2.7B}} & \multicolumn{1}{l|}{\textbf{OPT6.7B}} & \multicolumn{1}{l|}{\textbf{OPT13B}} & \multicolumn{1}{l|}{\textbf{OPT30B}} \\ \cline{2-13} 
\textbf{Recall} & \multicolumn{1}{r|}{0.3149} & \multicolumn{1}{r|}{0.3381} & \multicolumn{1}{r|}{0.3574} & 0.3007 & \multicolumn{1}{r|}{0.1933} & \multicolumn{1}{r|}{0.2429} & \multicolumn{1}{r|}{0.2128} & 0.203 & \multicolumn{1}{r|}{0.6077} & \multicolumn{1}{r|}{0.6142} & \multicolumn{1}{r|}{0.641} & \multicolumn{1}{r|}{0.6203} \\
\textbf{Precision} & \multicolumn{1}{r|}{0.6786} & \multicolumn{1}{r|}{0.7395} & \multicolumn{1}{r|}{0.7924} & 0.8163 & \multicolumn{1}{r|}{0.5737} & \multicolumn{1}{r|}{0.6375} & \multicolumn{1}{r|}{0.6944} & 0.7297 & \multicolumn{1}{r|}{0.6433} & \multicolumn{1}{r|}{0.6482} & \multicolumn{1}{r|}{0.6976} & \multicolumn{1}{r|}{0.7366} \\
\textbf{Accuracy} & \multicolumn{1}{r|}{0.698} & \multicolumn{1}{r|}{0.672} & \multicolumn{1}{r|}{0.654} & 0.592 & \multicolumn{1}{r|}{0.656} & \multicolumn{1}{r|}{0.624} & \multicolumn{1}{r|}{0.586} & 0.536 & \multicolumn{1}{r|}{0.7349} & \multicolumn{1}{r|}{0.6973} & \multicolumn{1}{r|}{0.7008} & \multicolumn{1}{r|}{0.6793} \\
\textbf{F1} & \multicolumn{1}{r|}{0.4301} & \multicolumn{1}{r|}{0.4641} & \multicolumn{1}{r|}{0.4927} & 0.4396 & \multicolumn{1}{r|}{0.2892} & \multicolumn{1}{r|}{0.3517} & \multicolumn{1}{r|}{0.3257} & 0.3176 & \multicolumn{1}{r|}{0.625} & \multicolumn{1}{r|}{0.6308} & \multicolumn{1}{r|}{0.6681} & \multicolumn{1}{r|}{0.6735} \\
 & \multicolumn{4}{c|}{\textbf{NLI-LM-2(Mis12B)}} & \multicolumn{4}{c|}{\textbf{NLI-LM-3(Mis7B)}} & \multicolumn{4}{c}{\textbf{LM-1-Step-Support(Mis7B)}} \\ \cline{2-13} 
\textbf{} & \multicolumn{1}{l|}{\textbf{OPT2.7B}} & \multicolumn{1}{l|}{\textbf{OPT6.7B}} & \multicolumn{1}{l|}{\textbf{OPT13B}} & \multicolumn{1}{l|}{\textbf{OPT30B}} & \multicolumn{1}{l|}{\textbf{OPT2.7B}} & \multicolumn{1}{l|}{\textbf{OPT6.7B}} & \multicolumn{1}{l|}{\textbf{OPT13B}} & \multicolumn{1}{l|}{\textbf{OPT30B}} & \multicolumn{1}{l|}{\textbf{OPT2.7B}} & \multicolumn{1}{l|}{\textbf{OPT6.7B}} & \multicolumn{1}{l|}{\textbf{OPT13B}} & \multicolumn{1}{l|}{\textbf{OPT30B}} \\ \cline{2-13} 
\textbf{Recall} & \multicolumn{1}{r|}{0.5635} & \multicolumn{1}{c|}{0.5810} & \multicolumn{1}{r|}{0.6340} & 0.5902 & \multicolumn{1}{r|}{0.5359} & \multicolumn{1}{r|}{0.4476} & \multicolumn{1}{r|}{0.5617} & 0.5602 & \multicolumn{1}{r|}{0.7403} & \multicolumn{1}{r|}{0.6844} & \multicolumn{1}{r|}{0.7991} & \multicolumn{1}{r|}{0.7293} \\
\textbf{Precision} & \multicolumn{1}{r|}{0.5397} & \multicolumn{1}{r|}{0.5951} & \multicolumn{1}{r|}{0.6834} & 0.7202 & \multicolumn{1}{r|}{0.4619} & \multicolumn{1}{r|}{0.4795} & \multicolumn{1}{r|}{0.6197} & 0.6898 & \multicolumn{1}{r|}{0.7089} & \multicolumn{1}{r|}{0.7305} & \multicolumn{1}{r|}{0.757} & \multicolumn{1}{r|}{0.7886} \\
\textbf{Accuracy} & \multicolumn{1}{r|}{0.668} & \multicolumn{1}{r|}{0.658} & \multicolumn{1}{r|}{0.69} & 0.66 & \multicolumn{1}{r|}{0.606} & \multicolumn{1}{r|}{0.564} & \multicolumn{1}{r|}{0.632} & 0.632 & \multicolumn{1}{r|}{0.7947} & \multicolumn{1}{r|}{0.7626} & \multicolumn{1}{r|}{0.7851} & \multicolumn{1}{r|}{0.7515} \\
\textbf{F1} & \multicolumn{1}{r|}{0.5513} & \multicolumn{1}{r|}{0.5880} & \multicolumn{1}{r|}{0.6578} & 0.6488 & \multicolumn{1}{r|}{0.4961} & \multicolumn{1}{r|}{0.4631} & \multicolumn{1}{r|}{0.5893} & 0.6183 & \multicolumn{1}{r|}{{\ul \textbf{0.7243}}} & \multicolumn{1}{r|}{0.7067} & \multicolumn{1}{r|}{{\ul \textbf{0.7775}}} & \multicolumn{1}{r|}{0.7578} \\
\textbf{} & \multicolumn{4}{c|}{\textbf{LM-1-Step-Support(Mis12B)}} & \multicolumn{4}{c|}{\textbf{LM-1-Step-Support(Mis22B)}} & \multicolumn{4}{c}{\textbf{LM-1-Step-Plausible(Mis7B)}} \\ \cline{2-13} 
\textbf{} & \multicolumn{1}{l|}{\textbf{OPT2.7B}} & \multicolumn{1}{l|}{\textbf{OPT6.7B}} & \multicolumn{1}{l|}{\textbf{OPT13B}} & \multicolumn{1}{l|}{\textbf{OPT30B}} & \multicolumn{1}{l|}{\textbf{OPT2.7B}} & \multicolumn{1}{l|}{\textbf{OPT6.7B}} & \multicolumn{1}{l|}{\textbf{OPT13B}} & \multicolumn{1}{l|}{\textbf{OPT30B}} & \multicolumn{1}{l|}{\textbf{OPT2.7B}} & \multicolumn{1}{l|}{\textbf{OPT6.7B}} & \multicolumn{1}{l|}{\textbf{OPT13B}} & \multicolumn{1}{l|}{\textbf{OPT30B}} \\ \cline{2-13} 
\textbf{Recall} & \multicolumn{1}{r|}{0.6575} & \multicolumn{1}{r|}{0.6429} & \multicolumn{1}{r|}{0.7319} & 0.7030 & \multicolumn{1}{r|}{0.5580} & \multicolumn{1}{r|}{0.5571} & \multicolumn{1}{r|}{0.6851} & 0.6466 & \multicolumn{1}{r|}{0.5667} & \multicolumn{1}{r|}{0.5167} & \multicolumn{1}{r|}{0.7845} & \multicolumn{1}{r|}{0.7283} \\
\textbf{Precision} & \multicolumn{1}{r|}{0.6959} & \multicolumn{1}{r|}{0.7542} & \multicolumn{1}{r|}{0.7818} & 0.7857 & \multicolumn{1}{r|}{0.7710} & \multicolumn{1}{r|}{0.7748} & \multicolumn{1}{r|}{0.7854} & 0.8113 & \multicolumn{1}{r|}{0.5484} & \multicolumn{1}{r|}{0.5538} & \multicolumn{1}{r|}{0.7339} & \multicolumn{1}{r|}{0.7751} \\
\textbf{Accuracy} & \multicolumn{1}{r|}{0.7720} & \multicolumn{1}{r|}{0.7620} & \multicolumn{1}{r|}{0.7780} & 0.7400 & \multicolumn{1}{r|}{0.7800} & \multicolumn{1}{r|}{0.7460} & \multicolumn{1}{r|}{0.7640} & 0.7320 & \multicolumn{1}{r|}{0.6707} & \multicolumn{1}{r|}{0.6179} & \multicolumn{1}{r|}{0.7652} & \multicolumn{1}{r|}{0.7414} \\
\textbf{F1} & \multicolumn{1}{r|}{0.6761} & \multicolumn{1}{r|}{0.6941} & \multicolumn{1}{r|}{0.7560} & 0.7421 & \multicolumn{1}{r|}{0.6474} & \multicolumn{1}{r|}{0.6482} & \multicolumn{1}{r|}{0.7318} & 0.7197 & \multicolumn{1}{r|}{0.5574} & \multicolumn{1}{r|}{0.5347} & \multicolumn{1}{r|}{0.7583} & \multicolumn{1}{r|}{0.7510} \\
\textbf{} & \multicolumn{4}{c|}{\textbf{LM-1-Step-Plausible(Mis12B)}} & \multicolumn{4}{c|}{\textbf{LM-1-Step-Plausible(Mis22B)}} & \multicolumn{4}{c}{\textbf{LM-1-Step-COT(Mis12B)}} \\ \cline{2-13} 
\textbf{} & \multicolumn{1}{l|}{\textbf{OPT2.7B}} & \multicolumn{1}{l|}{\textbf{OPT6.7B}} & \multicolumn{1}{l|}{\textbf{OPT13B}} & \multicolumn{1}{l|}{\textbf{OPT30B}} & \multicolumn{1}{l|}{\textbf{OPT2.7B}} & \multicolumn{1}{l|}{\textbf{OPT6.7B}} & \multicolumn{1}{l|}{\textbf{OPT13B}} & \multicolumn{1}{l|}{\textbf{OPT30B}} & \multicolumn{1}{l|}{\textbf{OPT2.7B}} & \multicolumn{1}{l|}{\textbf{OPT6.7B}} & \multicolumn{1}{l|}{\textbf{OPT13B}} & \multicolumn{1}{l|}{\textbf{OPT30B}} \\ \cline{2-13} 
\textbf{Recall} & \multicolumn{1}{r|}{0.7514} & \multicolumn{1}{r|}{0.7143} & \multicolumn{1}{r|}{0.7957} & 0.7970 & \multicolumn{1}{r|}{0.7127} & \multicolumn{1}{r|}{0.6714} & \multicolumn{1}{r|}{0.7660} & 0.7331 & \multicolumn{1}{r|}{0.5909} & \multicolumn{1}{r|}{0.6329} & \multicolumn{1}{r|}{0.6070} & \multicolumn{1}{r|}{0.6809} \\
\textbf{Precision} & \multicolumn{1}{r|}{0.6634} & \multicolumn{1}{r|}{0.7212} & \multicolumn{1}{r|}{0.7276} & 0.7626 & \multicolumn{1}{r|}{0.7127} & \multicolumn{1}{r|}{0.6980} & \multicolumn{1}{r|}{0.7200} & 0.7617 & \multicolumn{1}{r|}{0.4643} & \multicolumn{1}{r|}{0.5901} & \multicolumn{1}{r|}{0.5206} & \multicolumn{1}{r|}{0.6272} \\
\textbf{Accuracy} & \multicolumn{1}{r|}{0.7720} & \multicolumn{1}{r|}{0.7640} & \multicolumn{1}{r|}{0.7640} & 0.7600 & \multicolumn{1}{r|}{0.7920} & \multicolumn{1}{r|}{0.7400} & \multicolumn{1}{r|}{0.7500} & 0.7360 & \multicolumn{1}{r|}{0.6057} & \multicolumn{1}{r|}{0.6571} & \multicolumn{1}{r|}{0.5542} & \multicolumn{1}{r|}{0.6181} \\
\textbf{F1} & \multicolumn{1}{r|}{0.7047} & \multicolumn{1}{r|}{{\ul \textbf{0.7177}}} & \multicolumn{1}{r|}{0.7602} & {\ul \textbf{0.7794}} & \multicolumn{1}{r|}{0.7127} & \multicolumn{1}{r|}{0.6845} & \multicolumn{1}{r|}{0.7423} & 0.7471 & \multicolumn{1}{r|}{0.5200} & \multicolumn{1}{r|}{0.6107} & \multicolumn{1}{r|}{0.5605} & \multicolumn{1}{r|}{0.6530} \\
\textbf{} & \multicolumn{4}{c|}{\textbf{LM-2-Steps-FewShot(Mis7B)}} & \multicolumn{4}{c|}{\textbf{LM-2-Steps-Support(Mis7B)}} & \multicolumn{4}{c}{\textbf{LM-2-Steps-NoContradiction(Mis7B)}} \\ \cline{2-13} 
\textbf{} & \multicolumn{1}{l|}{\textbf{OPT2.7B}} & \multicolumn{1}{l|}{\textbf{OPT6.7B}} & \multicolumn{1}{l|}{\textbf{OPT13B}} & \multicolumn{1}{l|}{\textbf{OPT30B}} & \multicolumn{1}{l|}{\textbf{OPT2.7B}} & \multicolumn{1}{l|}{\textbf{OPT6.7B}} & \multicolumn{1}{l|}{\textbf{OPT13B}} & \multicolumn{1}{l|}{\textbf{OPT30B}} & \multicolumn{1}{l|}{\textbf{OPT2.7B}} & \multicolumn{1}{l|}{\textbf{OPT6.7B}} & \multicolumn{1}{l|}{\textbf{OPT13B}} & \multicolumn{1}{l|}{\textbf{OPT30B}} \\ \cline{2-13} 
\textbf{Recall} & \multicolumn{1}{r|}{0.5865} & \multicolumn{1}{r|}{0.6328} & \multicolumn{1}{r|}{0.6271} & 0.5961 & \multicolumn{1}{r|}{0.6358} & \multicolumn{1}{r|}{0.6231} & \multicolumn{1}{r|}{0.6888} & 0.6328 & \multicolumn{1}{r|}{0.6348} & \multicolumn{1}{r|}{0.6390} & \multicolumn{1}{r|}{0.7130} & \multicolumn{1}{r|}{0.6513} \\
\textbf{Precision} & \multicolumn{1}{r|}{0.6213} & \multicolumn{1}{r|}{0.6359} & \multicolumn{1}{r|}{0.6682} & 0.7451 & \multicolumn{1}{r|}{0.6321} & \multicolumn{1}{r|}{0.6683} & \multicolumn{1}{r|}{0.6981} & 0.7500 & \multicolumn{1}{r|}{0.5947} & \multicolumn{1}{r|}{0.6190} & \multicolumn{1}{r|}{0.6612} & \multicolumn{1}{r|}{0.7203} \\
\textbf{Accuracy} & \multicolumn{1}{r|}{0.7148} & \multicolumn{1}{r|}{0.6912} & \multicolumn{1}{r|}{0.6797} & 0.6791 & \multicolumn{1}{r|}{0.7402} & \multicolumn{1}{r|}{0.7113} & \multicolumn{1}{r|}{0.7175} & 0.6973 & \multicolumn{1}{r|}{0.7131} & \multicolumn{1}{r|}{0.6834} & \multicolumn{1}{r|}{0.6951} & \multicolumn{1}{r|}{0.6821} \\
\textbf{F1} & \multicolumn{1}{r|}{0.6034} & \multicolumn{1}{r|}{0.6343} & \multicolumn{1}{r|}{0.6471} & 0.6623 & \multicolumn{1}{r|}{0.6340} & \multicolumn{1}{r|}{0.6450} & \multicolumn{1}{r|}{0.6935} & 0.6864 & \multicolumn{1}{r|}{0.6141} & \multicolumn{1}{r|}{0.6288} & \multicolumn{1}{r|}{0.6861} & \multicolumn{1}{r|}{0.6841}
\end{tabular}%
}
\caption{Performance of various adequacy classifiers on the manual adequacy annotations from the random 50 contexts chosen from AbgCOQA.}
\end{table*}

Hugging Face\footnote{\url{https://huggingface.co}}
 was used to access the models (generators and classifiers). For all experiments A100 GPUs were used, with the exception of  OPT30B, for which we used H100 GPUS. The total GPU time for all experiments part of this paper were around 30 hours long.
\subsection{Manual annotations}
\label{appendix:manual_annotations}
We randomly sample 50 context-ambiguous question pairs from Abg-COQA’s training set and manually annotate for the semantic equivalence and adequacy of the sampled responses and the greedy response, for all OPT models (2.7B-30B). 
For adequacy, we annotate using finer-grained notions, as seen in \Cref{table:correctness_labels} (these are also the options given to the annotators when completing the labelling task), which are ultimately mapped to a binary adequacy decision — used as correctness labels during evaluation.
For the manual annotations regarding Provo Corpus, we randomly choose a subset of 5 contexts from the original 100 contexts we sampled from the dataset. We annotate the OPT models' generations for the 5 contexts and the corresponding generations for the 5 corrupted contexts as `Plausible' or `Not plausible' given the original contexts. Hence, we annotate 100 generations in total (50 from the original and 50 from the corrupted contexts).

The authors of this paper were the ones who undertook the task of annotation. For each assessment, 1 annotator hand labelled each item. 
Specific instructions given for each task:

\paragraph{Semantic Equivalence.}
`A passage and a question (given the passage) are provided, which you need to read. Two answers to the question are also provided. You need to decide if these two answers mean the same thing (given the question), or not.’

\paragraph{Adequacy to the prompt (QA).}
`A passage, a question (given the passage) and some possible answers are provided, which you need to read. You are given 1 possible answer to the prompt. You need to decide whether this answer is adequate/plausible to the prompt (`Match (fully) 1 plausible answer'), with respect to the question, the passage and the possible answers, or not (`Wrong', or `Inability to answer' --- the response explains how the model can not provide an answer, \eg `I don't know.'). If you think the answer is still plausible given the passage and question, but not included in the possible answers, you might still assign it as plausible, given the available option (`Plausible but not in references'). The same goes for answers that might include more than one (or all) of answers you deem plausible, given the references and your own judgement (options `Multiple plausible answers found' and `All plausible answers found'). If you think the question partly includes a correct answer (e.g. part of the reference is in the response, or the full reference answer as well as some additional information), you can choose the relevant option (`Match (partly) 1 plausible answer').’

\paragraph{Adequacy to the prompt (NWP).}
`You are given a context and a plausible continuation. You need to decide whether this continuation is plausible or not, given the context. By plausible, we mean that when appending the continuation to the context, the new piece of text created remains sensical, comprehensible and grammatically correct.'

\begin{table}[t]
    \centering
    \begin{tabular}{lll}
        \toprule
        Fine-grained label & Binary  \\
        &  decision \\
        \midrule
        Inability to answer &  Incorrect \\
        Wrong &  Incorrect \\
        Match (fully) 1 plausible answer & Correct \\
        Match (partly) 1 plausible answer & Correct \\
        Multiple plausible answers found & Correct \\
        All plausible answers found & Correct \\
        Plausible but not in references & Correct \\
        \bottomrule
    \end{tabular}
\caption{Fine-grained correctness labels and their associate binary correctness assigned when computing evaluation metrics.}
\label{table:correctness_labels}
\end{table}

\section{Results}
\subsection{Evaluating adequacy classifiers}
\label{appendix:eval_adeq_classifiers}
The manual annotations are used to evaluate the performance of the various classifiers used. For each analysed model (OPT2.7B-30B), we have annotations for each sampled response for all contexts we annotated (50x10 = 500 annotations, for each model). We compare the predictions from the various classifiers discussed in \Cref{appendix:adequacy_classifiers} to the manual annotations. Various performance statistics (accuracy, precision, recall and F1-score) are measured. We decide on which classifier to employ for experiments based on F1-score. The two best performing models are \texttt{LM-1-Step-Support-Mis7B} and \texttt{LM-1-Step-Plausible-Mis12B}, with the former achieving the best performance for OPT2.7B and OPT13B and the latter achieving the best performance for OPT6.7B and OPT30B; we employ the latter in our experiments. 
Similar results can be seen in \Cref{tab:eval_nwp_classifier} for the adequacy classifier for the NWP task (given the manual annotations discussed in \Cref{appendix:manual_annotations}).

\begin{table}[]
\centering
\begin{tabular}{lrrr}
\multicolumn{4}{c}{\textbf{Adequacy Classifier Performance}} \\ \hline
\multicolumn{1}{l|}{} & \multicolumn{1}{c}{\textbf{Original}} & \multicolumn{1}{c}{\textbf{Corrupt}} & \multicolumn{1}{c}{\textbf{Total}} \\ \hline
 
\multicolumn{1}{l|}{\textbf{Recall}} & 0.936 & 1 & 0.958 \\
 
\multicolumn{1}{l|}{\textbf{Precision}} & 0.936 & 0.556 & 0.75 \\
 
\multicolumn{1}{l|}{\textbf{Accuracy}} & 0.88 & 0.6 & 0.74 \\
 
\multicolumn{1}{l|}{\textbf{F1}} & 0.936 & 0.714 & 0.841
\end{tabular}
\caption{Performance of adequacy classifier for Provo Corpus based on the manually annotated continuations given the original contexts, corrupted contexts and their aggregation.}
\label{tab:eval_nwp_classifier}
\end{table}

\subsection{Evaluating AUROC correctness decision}
\label{appendix:eval_auroc_correct}

When computing AUROC, the correctness decision for the greedy response needs to be automated. We assess various ways to achieve this. \citet{kuhn2022semantic} uses Rouge-L score \cite{lin2004automatic} between the model's response and a reference. If this value surpasses a threshold (set by \citeauthor{kuhn2022semantic} to 0.3), then the response is considered correct (or otherwise, adequate).
However, during our analysis, we conclude that the performance of this heuristic can be rather impactful on the AUROC values. If the heuristic misclassifies a response's correctness, a more informative confidence metric's AUROC will be negatively impacted to a higher extent than a less informative metric. For this reason, we find it important to evaluate the various heuristics for assessing a response's correctness and evaluating their performance. For this purpose, we exploit the manual annotations for the adequacy of the greedy response as gold-labels, which we compare against the automated correctness/adequacy decision from different methods (for each analysed model OPT2.7B-30B, we have 50 annotations for the adequacy of the greedy). We assess \citeauthor{kuhn2022semantic} automated correctness and we also assess different LMs as a judge, similar to \citet{lin2024generating}. We employ \texttt{Mistral-Small-Instruct-2409} and \texttt{gpt3.5-turbo}, referred to as \texttt{Mis22B} and \texttt{ChatGPT}). The prompt includes the passage (if relevant), the question, the reference responses and the `proposed' response (\emph{i.e.} the greedy response). Find below the exact prompts: 

\paragraph{Prompt for AbgCOQA.}

\texttt{You are presented with a document, a question based on the document, some acceptable answers and a proposed answer. Generate True if the proposed answer is a plausible answer to the question given the document and False if not. By plausible, I mean that the answer might be conveying the same meaning as one of the acceptable answers (even if it contains more or less information, as long as they have the same meaning), or, even if not similar to one of the acceptable answers, the answer can still be supported by the document.}

\texttt{Document:'<PASSAGE>'}

\texttt{Question:'<AMBIGUOUS\_QUESTION>'}

\texttt{Acceptable Answers: '<REFERENCES>'}

\texttt{Proposed Answer: '<GREEDY>'.}

\paragraph{Prompt for AmbigQA.}

\texttt{You are presented with a question, some acceptable answers and a proposed answer. Generate True if the proposed answer is a plausible answer to the question given your training data and False if not. By plausible, I mean that the answer might be conveying the same meaning as one of the acceptable answers (even if it contains more or less information, as long as they have the same meaning), or, even if not similar to one of the acceptable answers, the answer can still be supported by your training data.}

\texttt{Question:'<AMBIGUOUS\_QUESTION>'}

\texttt{Acceptable Answers: '<REFERENCES>'}

\texttt{Proposed Answer: '<GREEDY>'.}

\begin{table}
\begin{tabular}{l|ccc}
\textbf{} & \multicolumn{1}{l}{\textbf{RougeL}} & \multicolumn{1}{l}{\textbf{Mis22B}} & \multicolumn{1}{l}{\textbf{ChatGPT}} \\ \hline
\textbf{} & \multicolumn{3}{c}{\textbf{OPT2.7B}} \\ \hline
\textbf{Recall} & 0.767 & 0.7667 & 0.9 \\
\textbf{Precision} & 0.885 & 0.821 & 0.844 \\
\textbf{Accuracy} & 0.8 & 0.76 & 0.84 \\
\textbf{F1} & 0.821 & 0.793 & \textbf{0.871} \\ \hline
 & \multicolumn{3}{c}{\textbf{OPT6.7B}} \\ \hline
\textbf{Recall} & 0.694 & 0.75 & 0.861 \\
\textbf{Precision} & 0.893 & 0.931 & 0.912 \\
\textbf{Accuracy} & 0.72 & 0.78 & 0.84 \\
\textbf{F1} & 0.781 & 0.831 & \textbf{0.886} \\ \hline
 & \multicolumn{3}{c}{\textbf{OPT13B}} \\ \hline
\textbf{Recall} & 0.730 & 0.838 & 0.946 \\
\textbf{Precision} & 0.931 & 0.886 & 0.921 \\
\textbf{Accuracy} & 0.76 & 0.8 & 0.9 \\
\textbf{F1} & 0.812 & 0.861 & \textbf{0.933} \\ \hline
 & \multicolumn{3}{c}{\textbf{OPT30B}} \\ \hline
\textbf{Recall} & 0.706 & 0.824 & 0.882 \\
\textbf{Precision} & 0.923 & {\color[HTML]{1F1F1F} 0.903} & {\color[HTML]{1F1F1F} 0.909} \\
\textbf{Accuracy} & 0.76 & {\color[HTML]{1F1F1F} 0.82} & {\color[HTML]{1F1F1F} 0.86} \\
\textbf{F1} & 0.8 & {\color[HTML]{1F1F1F} 0.862} & {\color[HTML]{1F1F1F} \textbf{0.896}}
\end{tabular}
\caption{Performance of various automated correctness decision methods: RougeL(response, reference) > 0.3 (RougeL), Mis22B as a judge and ChatGPT as a judge. ChatGPT outperforms other methods as per F1 (and other metrics), and is thus employed as the correctness criterion for QA experiments.}
\label{tab:automated_correctness_eval}
\end{table}

The results of this evaluation for AbgCOQA can be found in \Cref{tab:automated_correctness_eval}. Based on F1, we decide to employ ChatGPT as a judge as the automated correctness criterion for the QA tasks, as it outperforms other methods in our analysis by a large margin.

\begin{figure}
\includegraphics[width=6.5cm]{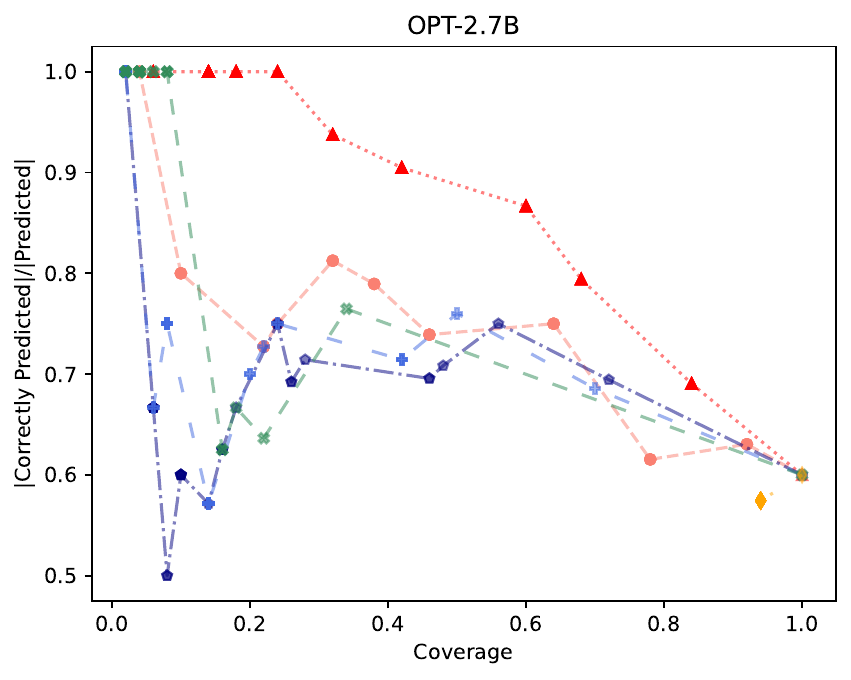}
\hfill
\includegraphics[width=6.5cm]{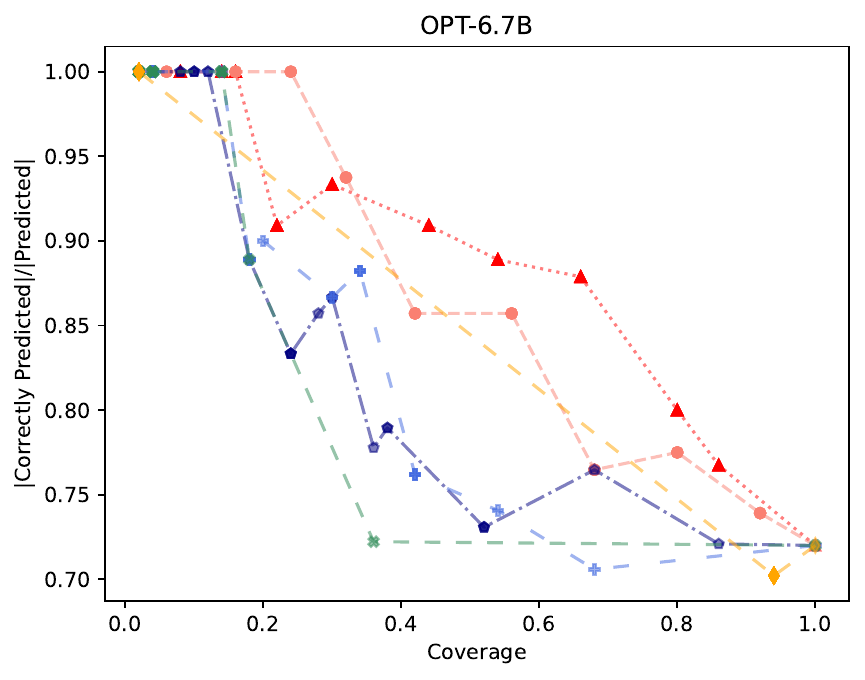}
\hfill
\includegraphics[width=6.5cm]{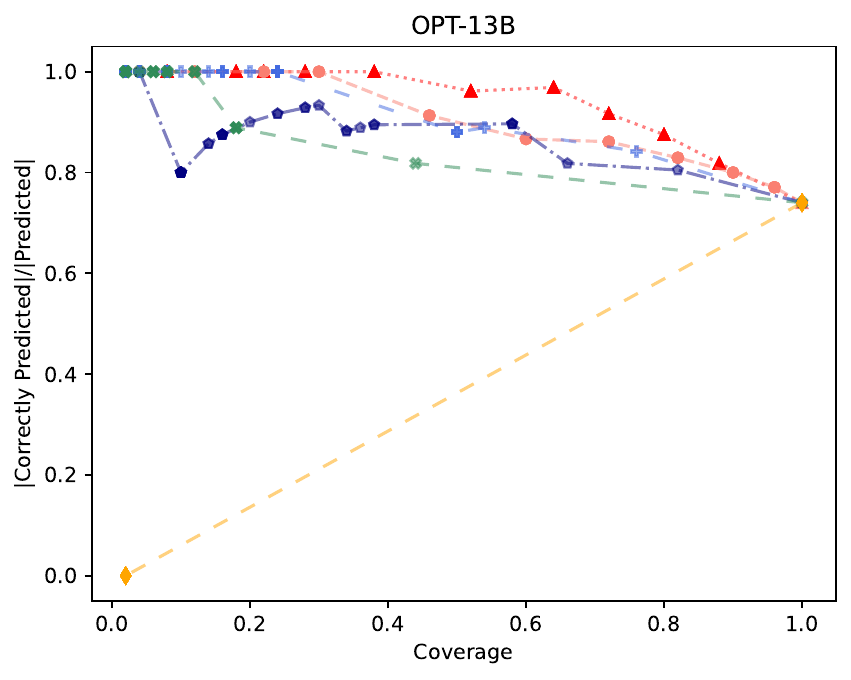}
\hfill
\includegraphics[width=6.5cm]{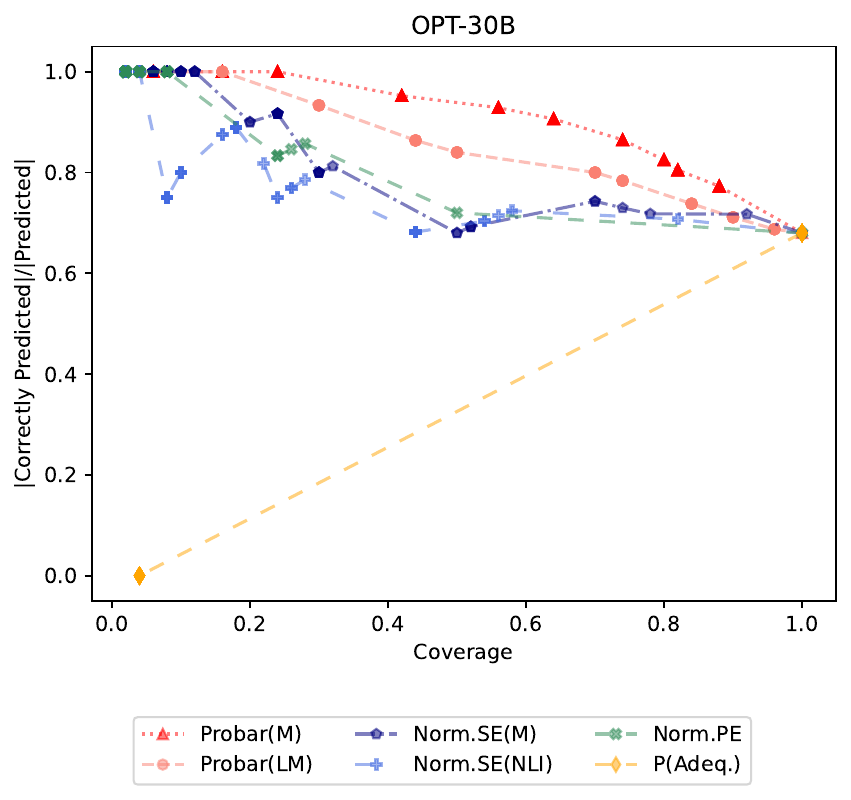}
\hfill
\caption{Coverage vs `Precision' for different uncertainty indicators, across OPT models (for manually annotated contexts).}
\label{fig:selective_gen_manual_annotations}
\end{figure}

\begin{figure}
\includegraphics[width=6.5cm]{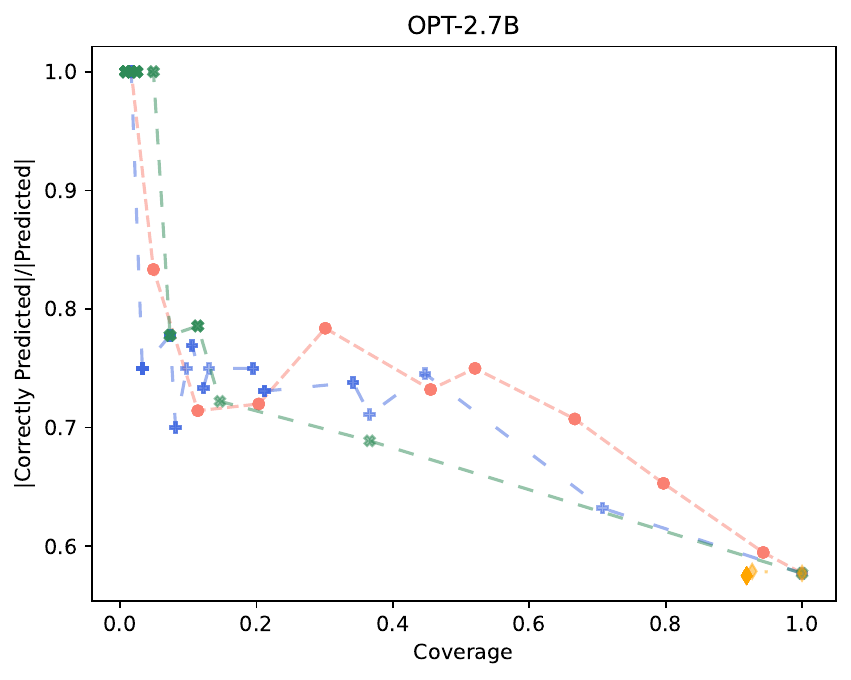}
\hfill
\includegraphics[width=6.5cm]{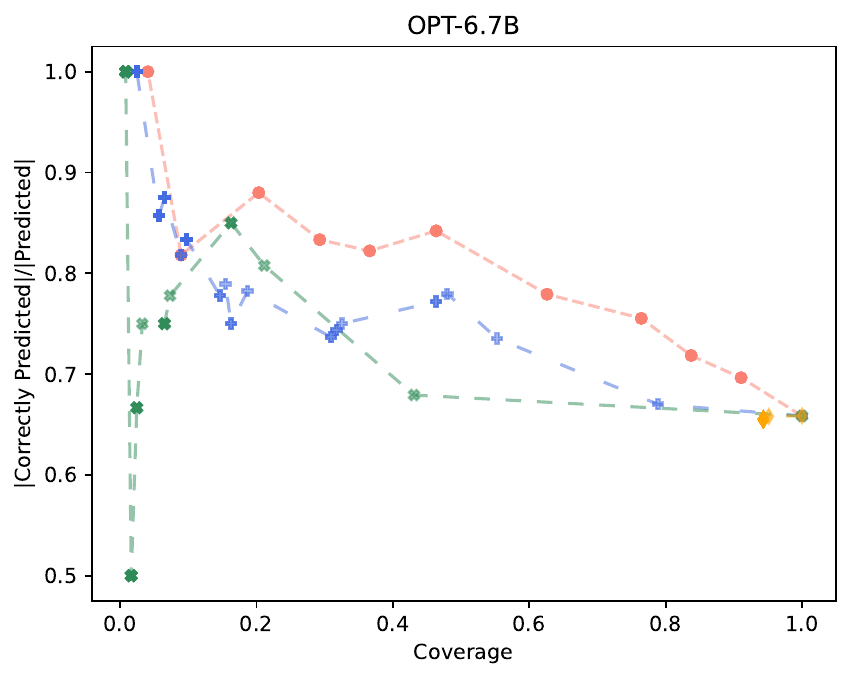}
\hfill
\includegraphics[width=6.5cm]{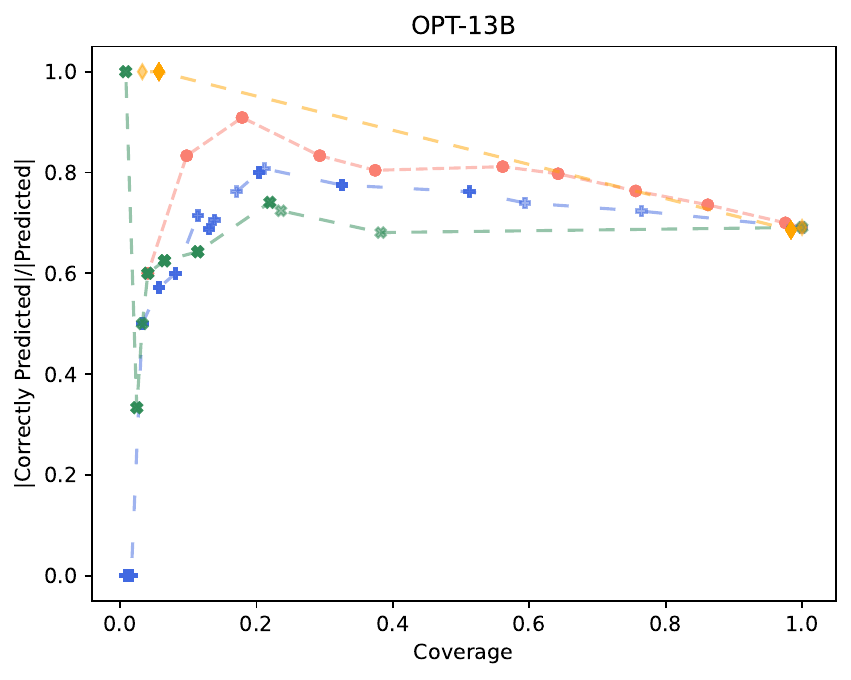}
\hfill
\includegraphics[width=6.5cm]{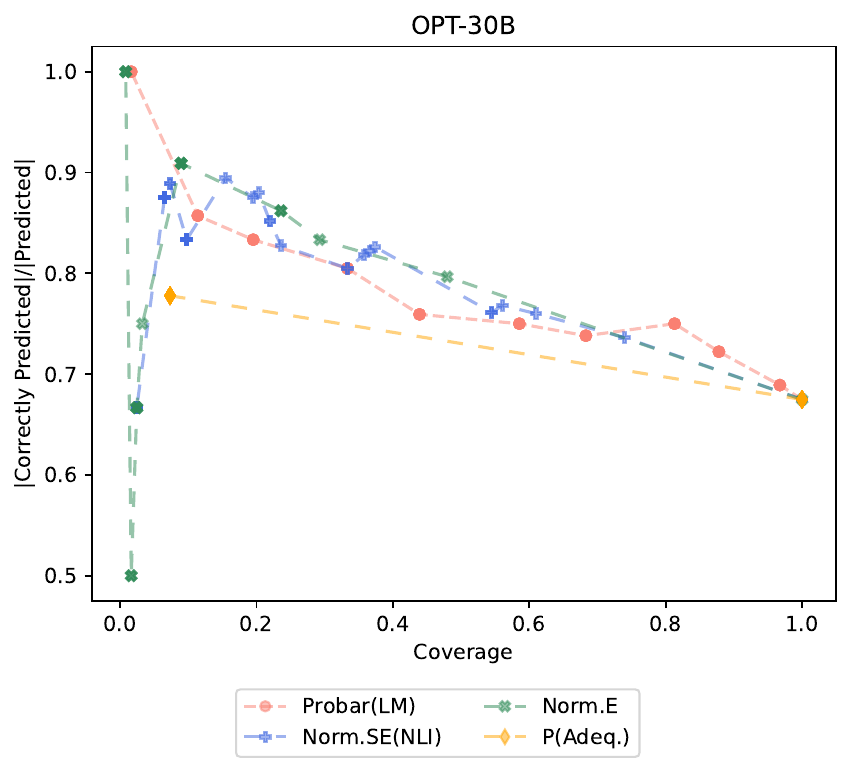}
\hfill
\caption{Coverage vs `Precision' for different uncertainty indicators, across OPT models (for ambiguous contexts from AbgCOQA's test set).}
\label{fig:selective_gen_ambig_test}
\end{figure}

\begin{figure}
\includegraphics[width=6.5cm]{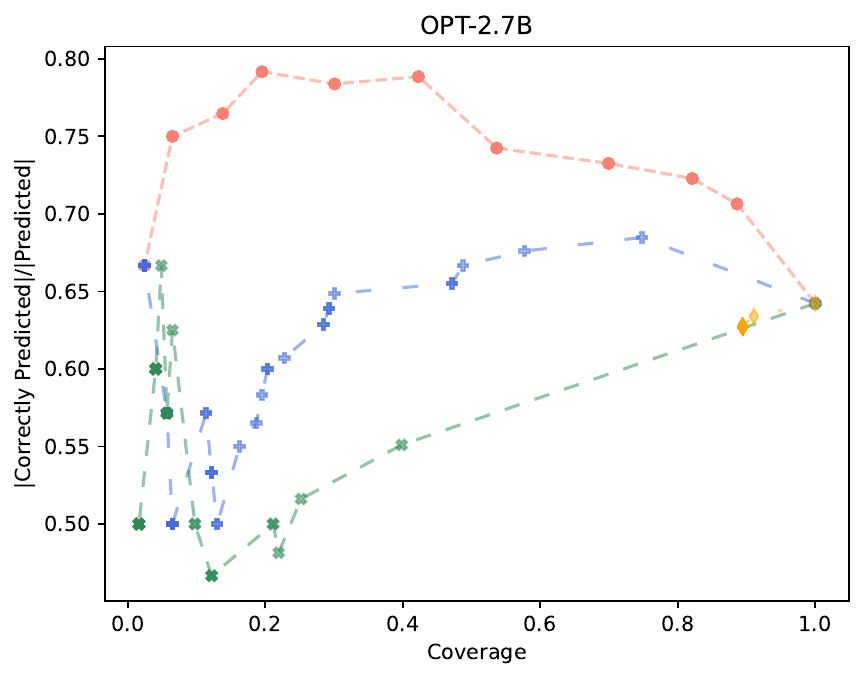}
\hfill
\includegraphics[width=6.5cm]{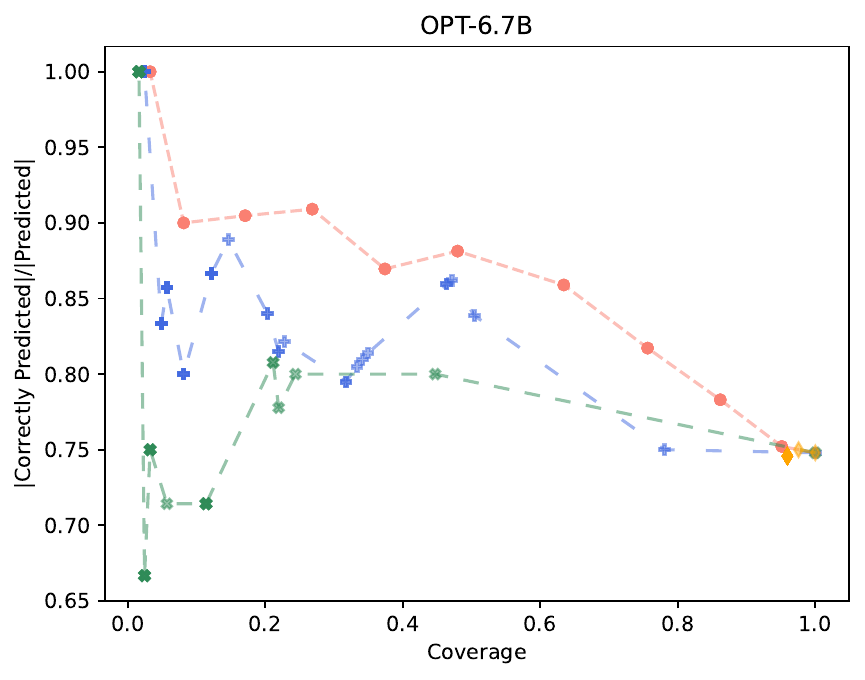}
\hfill
\includegraphics[width=6.5cm]{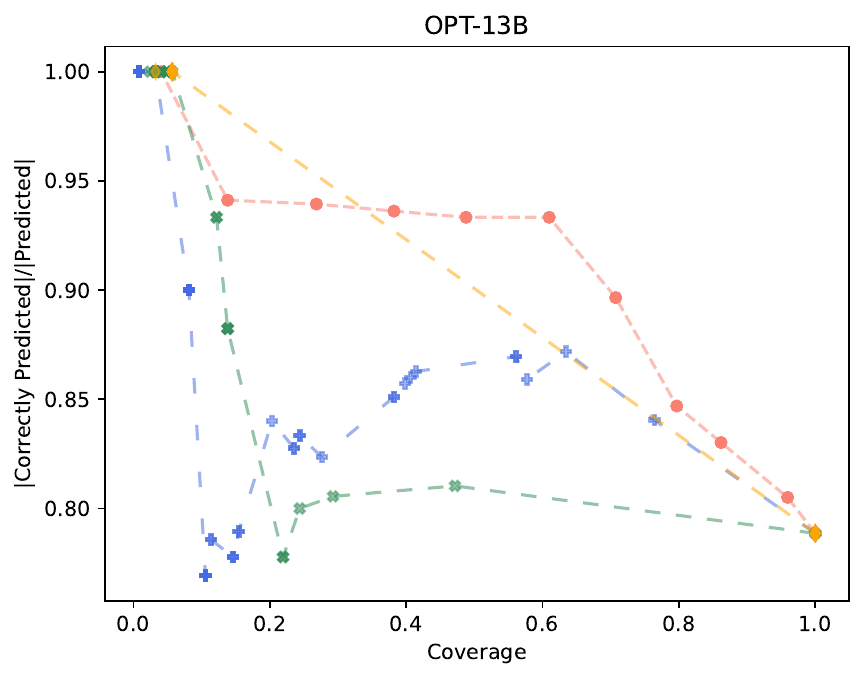}
\hfill
\includegraphics[width=6.5cm]{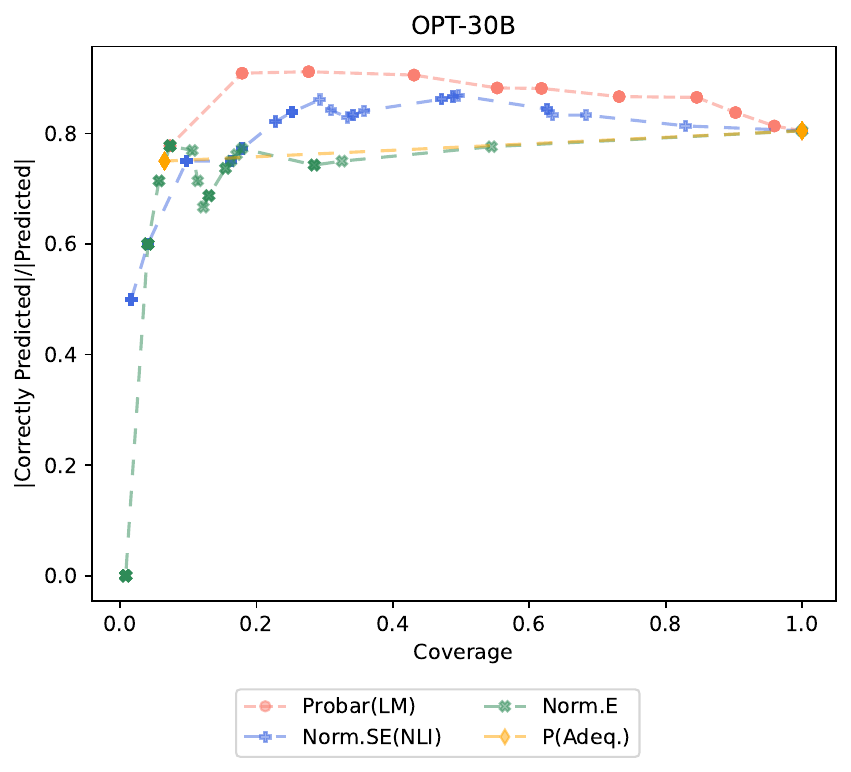}
\hfill
\caption{Coverage vs `Precision' for different uncertainty indicators, across OPT models (for non-ambiguous contexts from AbgCOQA's test set).}
\label{fig:selective_gen_non_ambig_test}

\end{figure}

\begin{figure}
\includegraphics[width=6.5cm]{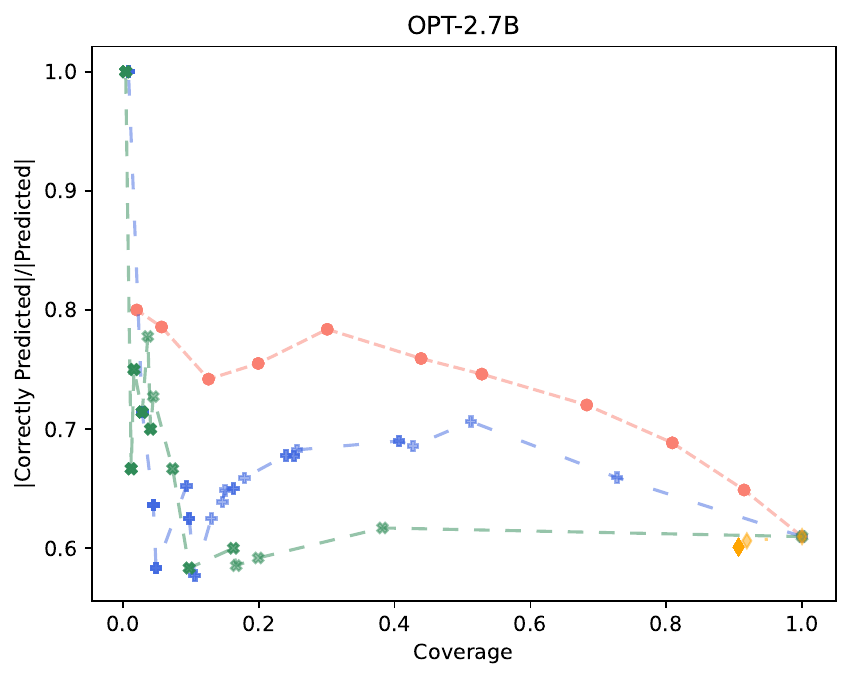}
\hfill
\includegraphics[width=6.5cm]{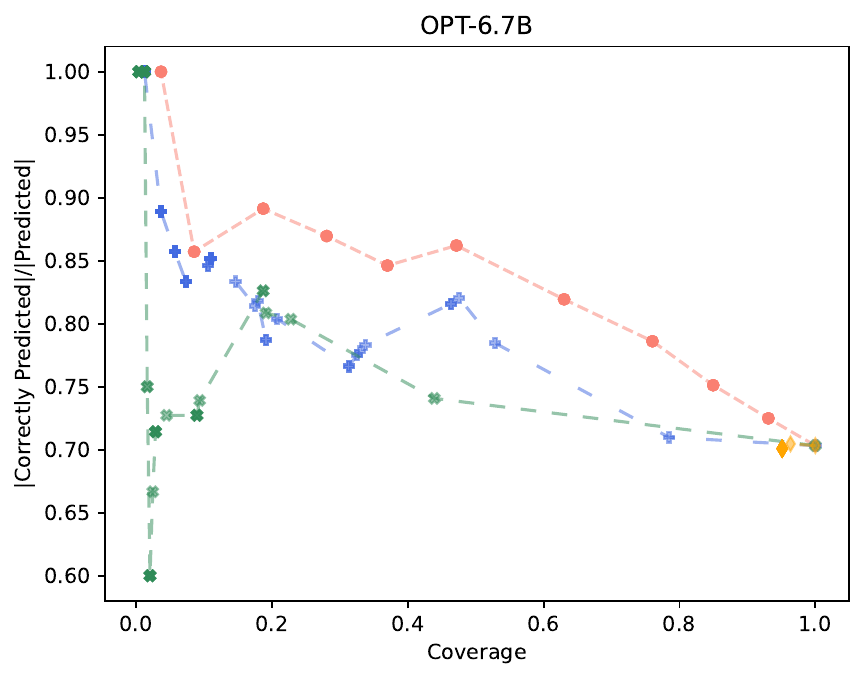}
\hfill
\includegraphics[width=6.5cm]{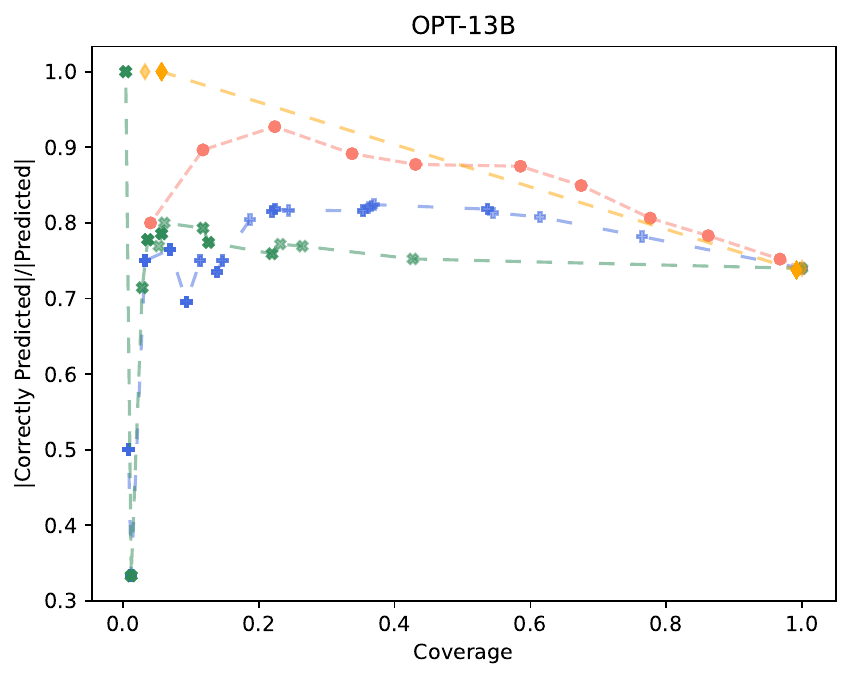}
\hfill
\includegraphics[width=6.5cm]{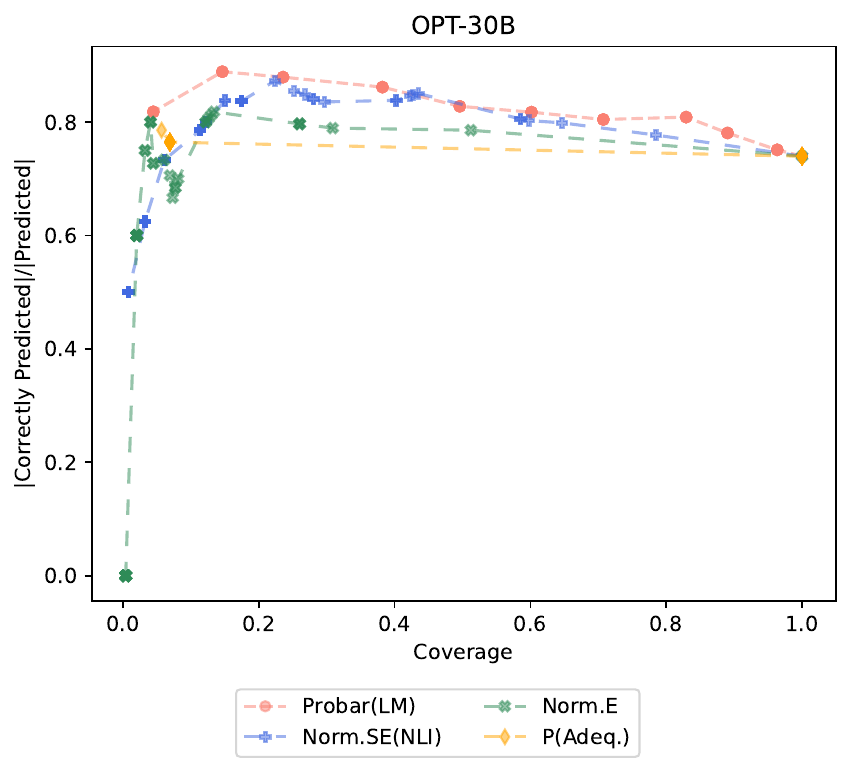}
\hfill
\caption{Coverage vs `Precision' for different uncertainty indicators, across OPT models (for mixed contexts from AbgCOQA's test set).}
\label{fig:selective_gen_mixed_test}

\end{figure}

\begin{figure}
\includegraphics[width=6.5cm]{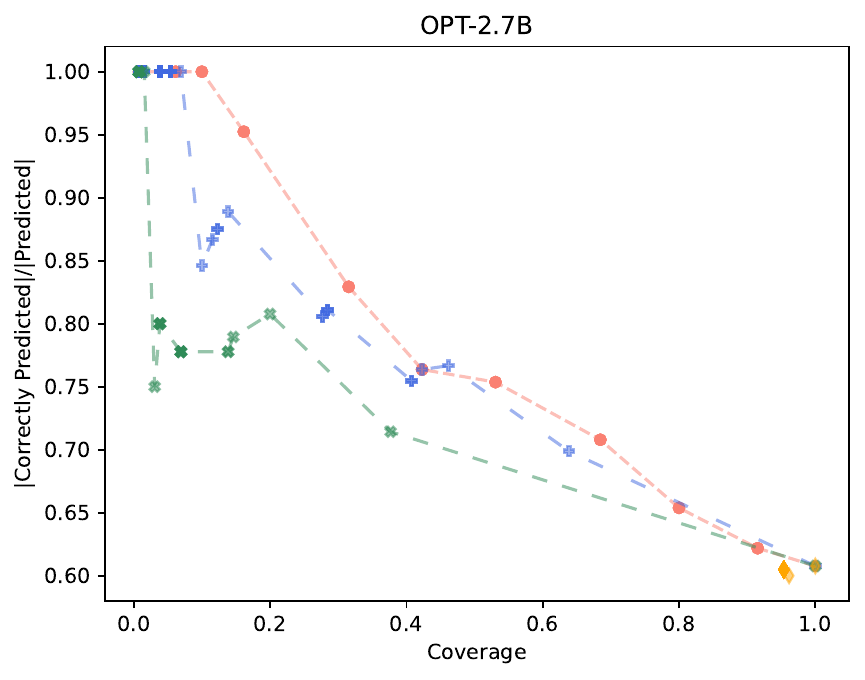}
\hfill
\includegraphics[width=6.5cm]{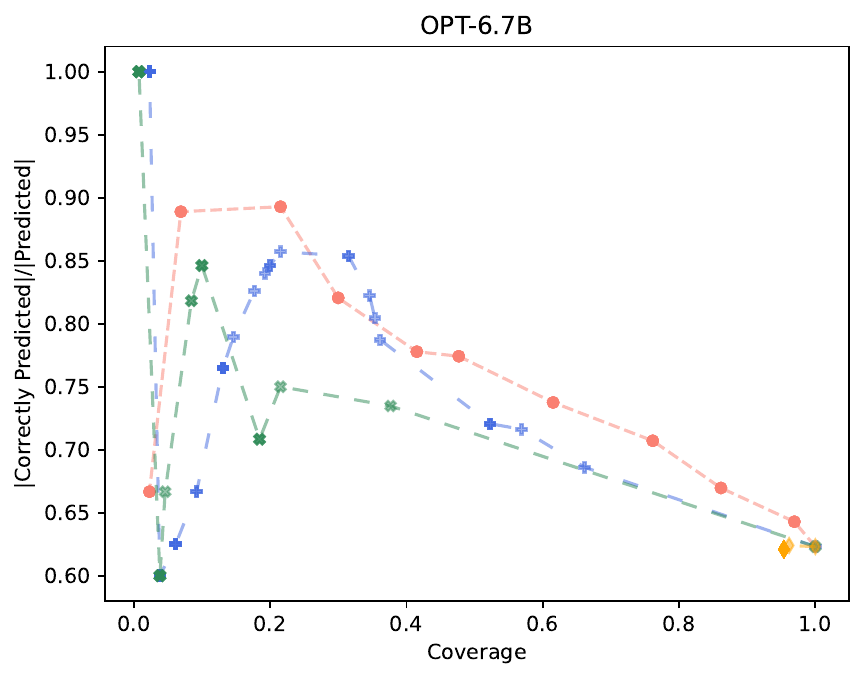}
\hfill
\includegraphics[width=6.5cm]{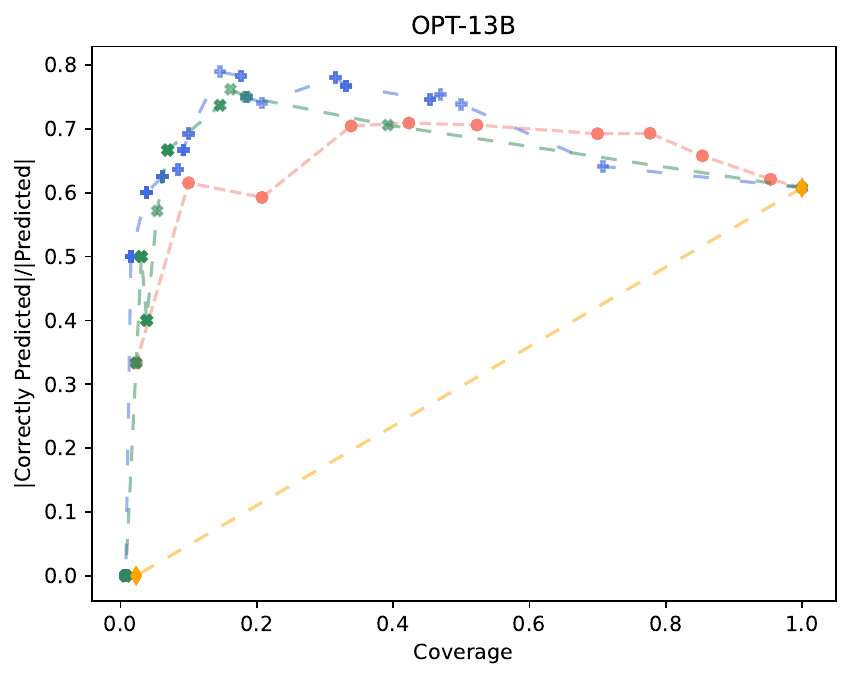}
\hfill
\includegraphics[width=6.5cm]{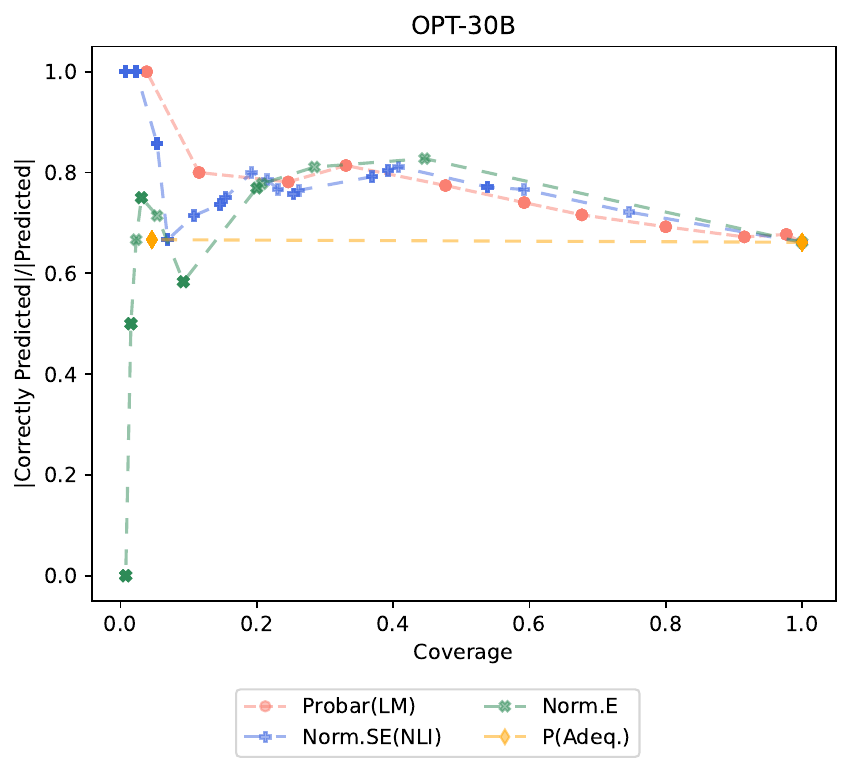}
\hfill
\caption{Coverage vs `Precision' for different uncertainty indicators, across OPT models (for ambiguous contexts from AbgCOQA's development set).}
\label{fig:selective_gen_ambig_dev}
\end{figure}

\begin{figure}
\includegraphics[width=6.5cm]{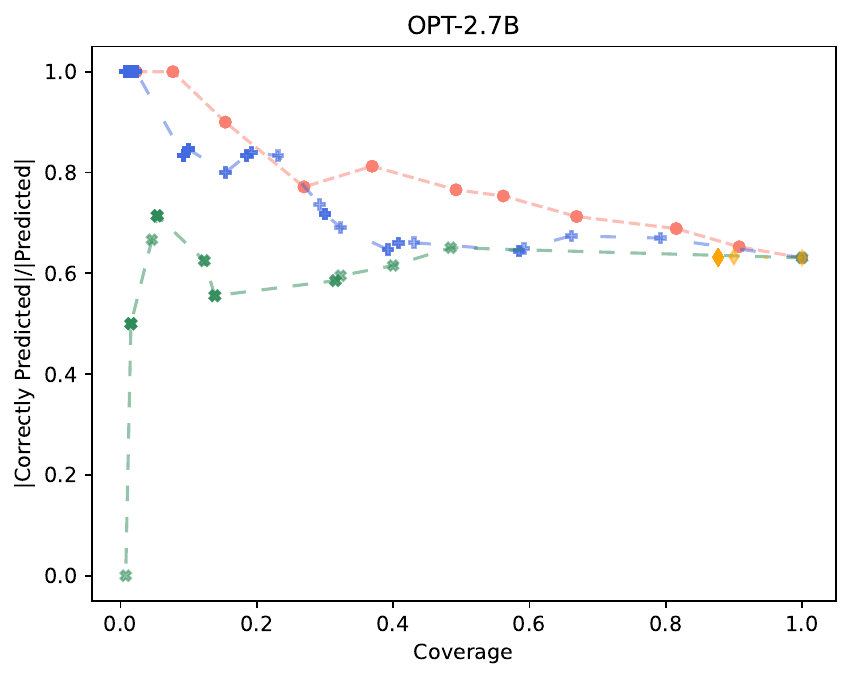}
\hfill
\includegraphics[width=6.5cm]{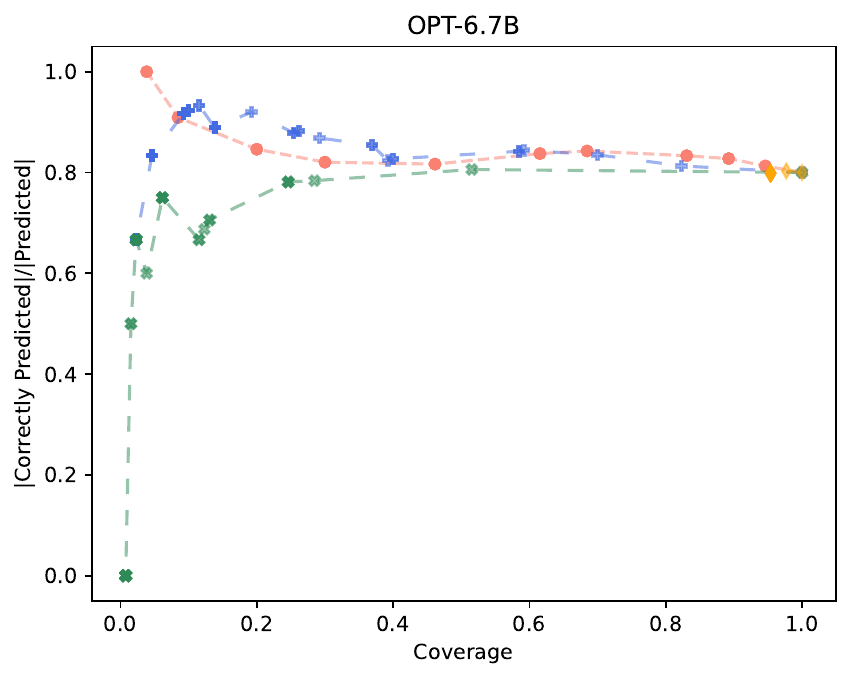}
\hfill
\includegraphics[width=6.5cm]{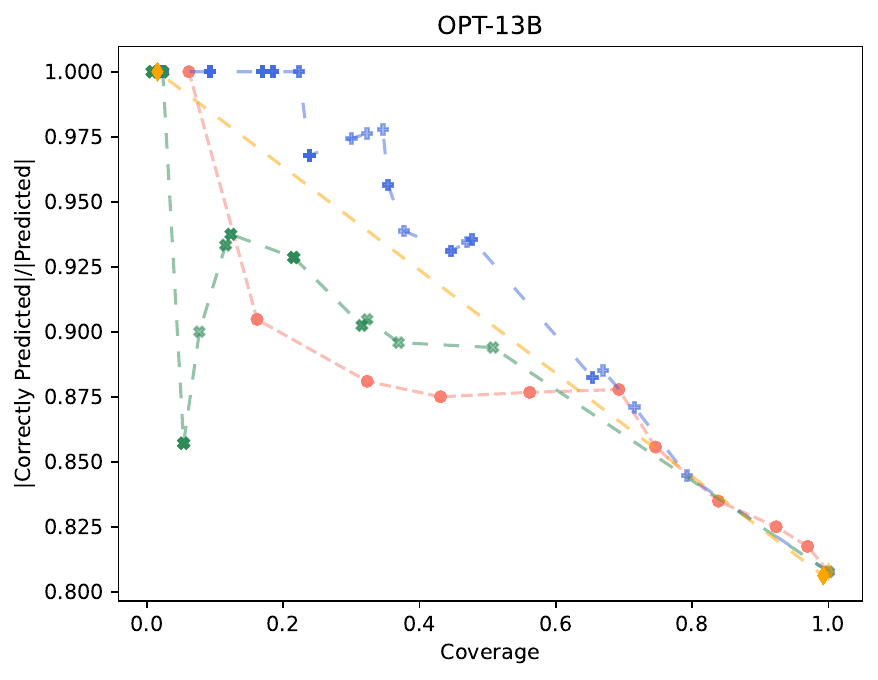}
\hfill
\includegraphics[width=6.5cm]{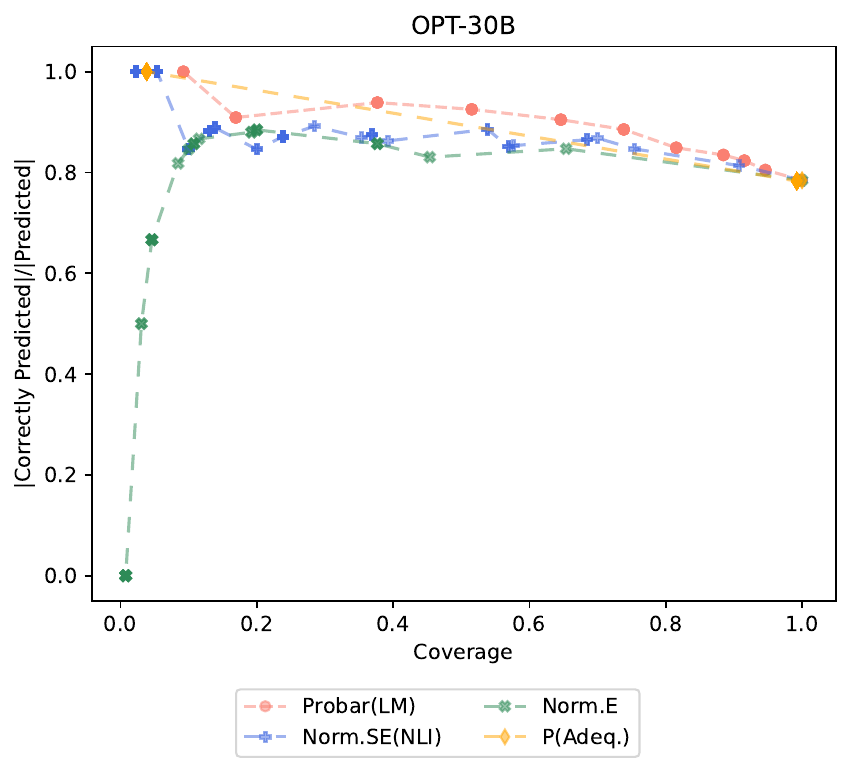}
\hfill
\caption{Coverage vs `Precision' for different uncertainty indicators, across OPT models (for non-ambiguous contexts from AbgCOQA's development set).}
\label{fig:selective_gen_non_ambig_dev}

\end{figure}

\begin{figure}
\includegraphics[width=6.5cm]{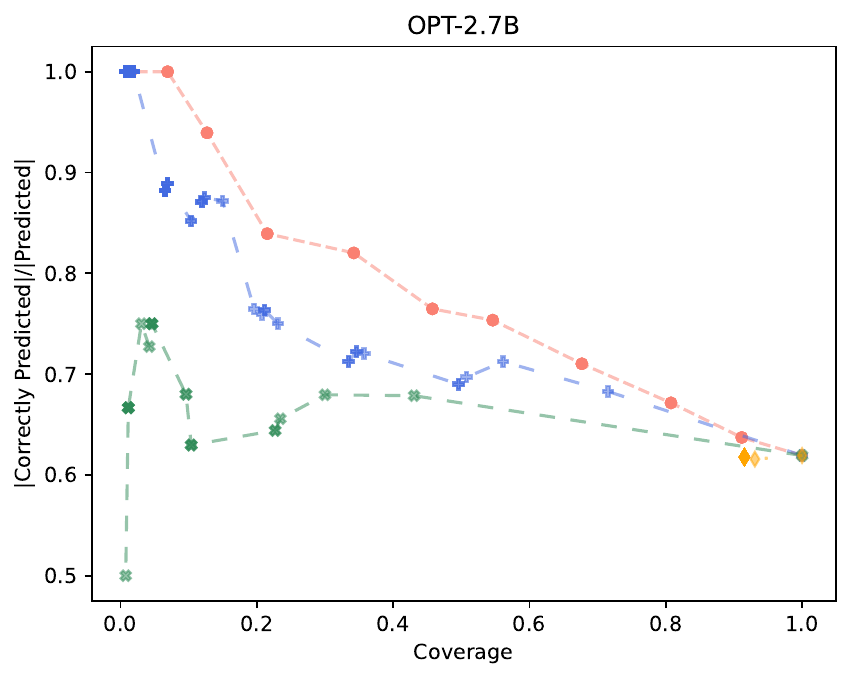}
\hfill
\includegraphics[width=6.5cm]{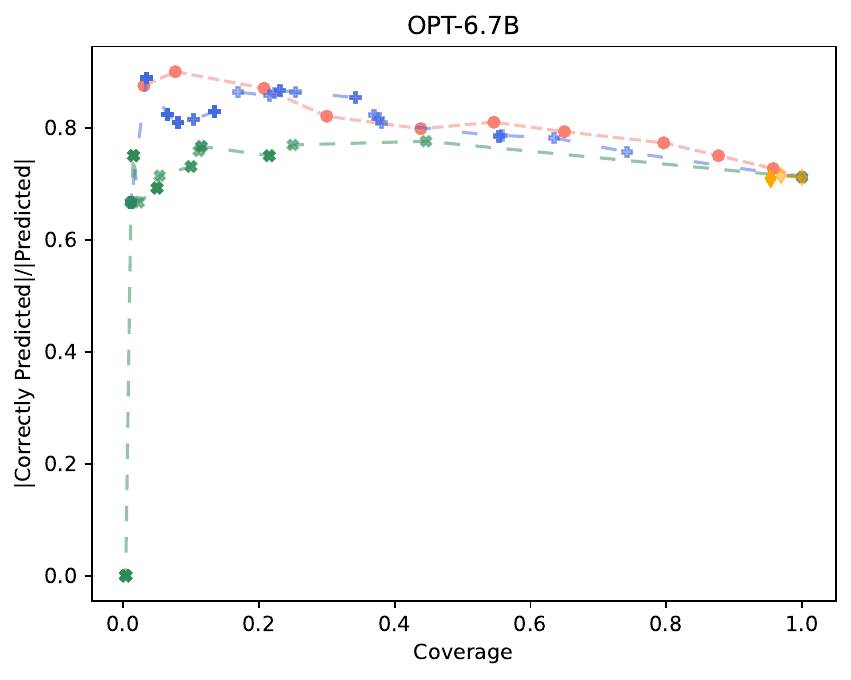}
\hfill
\includegraphics[width=6.5cm]{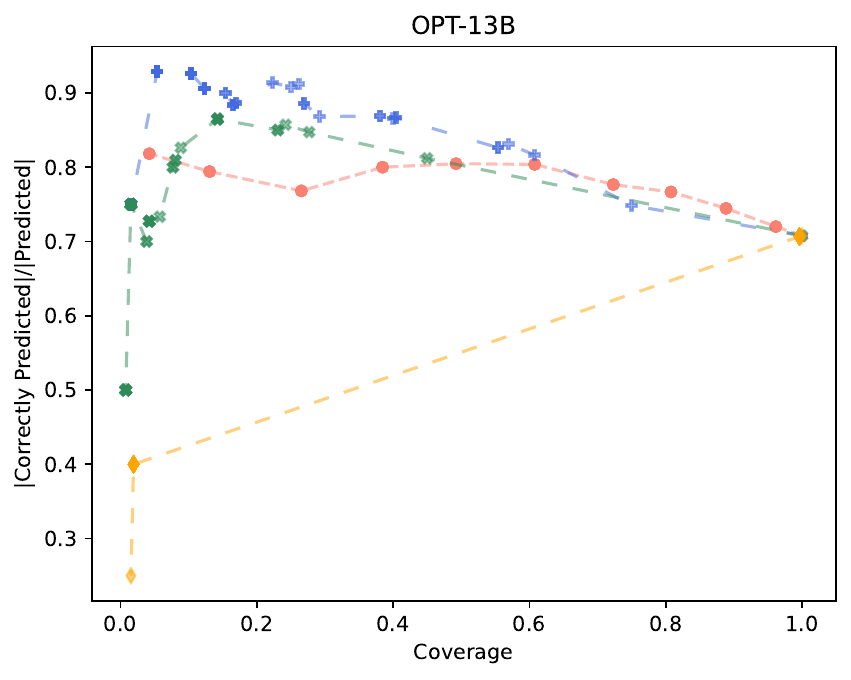}
\hfill
\includegraphics[width=6.5cm]{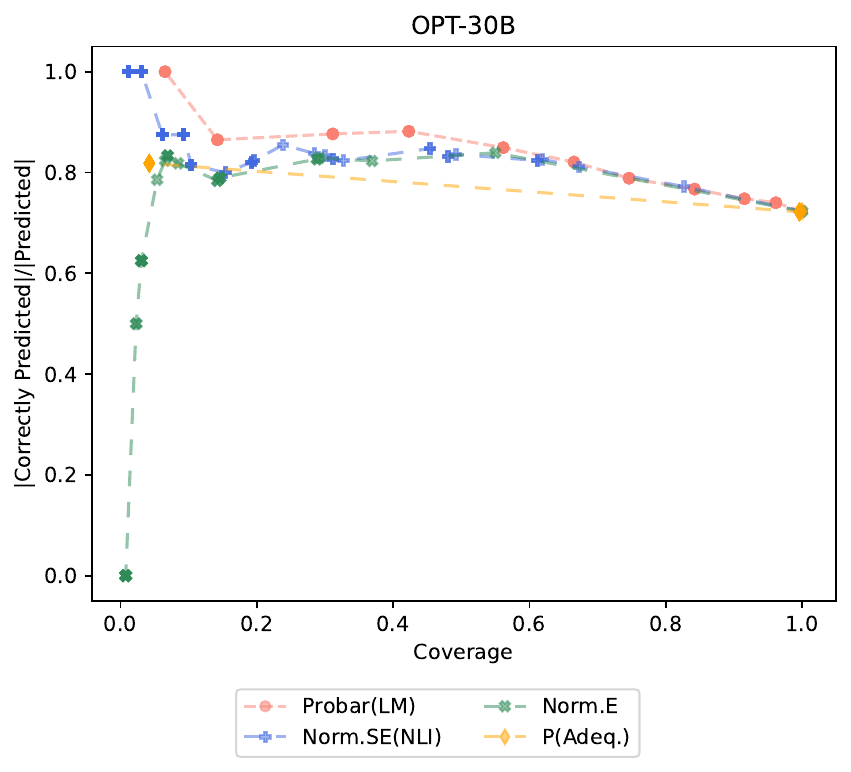}
\hfill
\caption{Coverage vs `Precision' for different uncertainty indicators, across OPT models (for mixed contexts from AbgCOQA's development set).}
\label{fig:selective_gen_mixed_dev}

\end{figure}

\subsection{Risk vs Coverage plots}
\label{appendix:risk_cov_plots}
Beyond the use of AUROC as an evaluation metric, we also plot the Selective Coverage vs Precision plots, another selective prediction setting in which one can assess the informativeness of a metric. The resulting plots can be found in \Cref{fig:selective_gen_manual_annotations}-\ref{fig:selective_gen_mixed_dev}. Across almost all settings, \probar outperforms other quantifiers.

\subsection{Additional Analysis}
\begin{figure*}[h!]
\centering
\includegraphics[width=15cm]{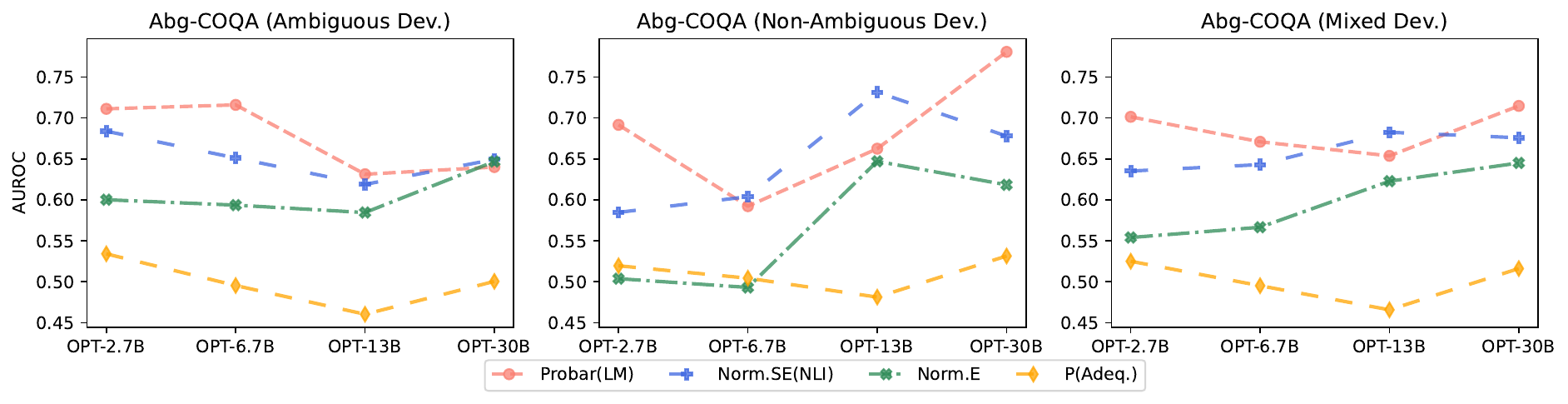}
\hfill
\caption{AUROC values for AbgCOQA's development set on ambiguous prompts (left), unambiguous prompts (middle) and their aggregation (right), where correctness of the greedy was assessed by \texttt{gpt3.5-turbo} as a judge. It concerns 130,130 and 260 prompts respectively.}
\label{fig:abgcoqa_dev_auroc_vs_model}
\end{figure*}

We also demonstrate in \Cref{fig:abgcoqa_dev_auroc_vs_model} the AUROC results for Abg-COQA's development set. For all models except OPT13B, \probar outperforms or performs on par with other baselines. 

\begin{figure*}[!h]
\centering
\begin{subfigure}[t]{6.5cm}
    \includegraphics[width=6.5cm]{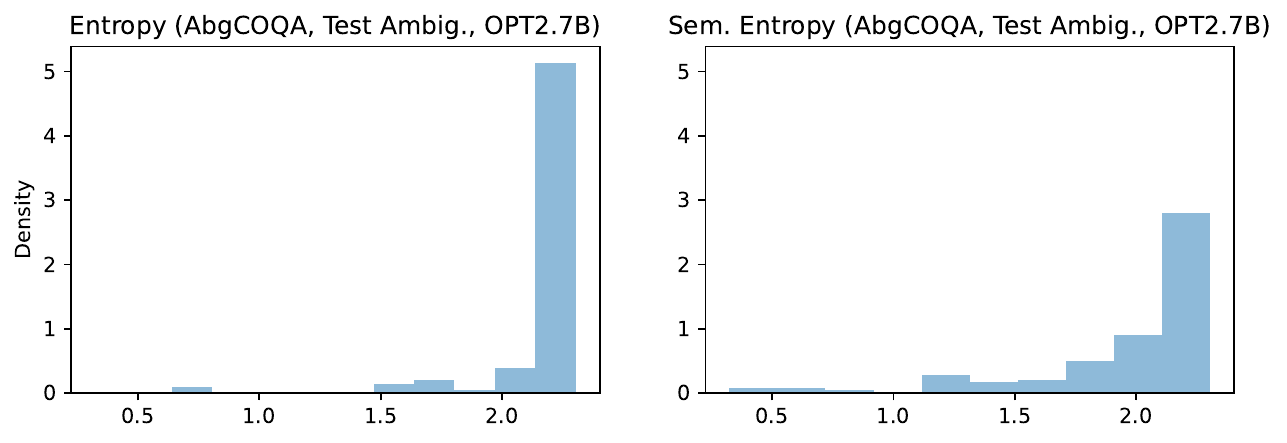}
    \hfill
    \includegraphics[width=6.5cm]{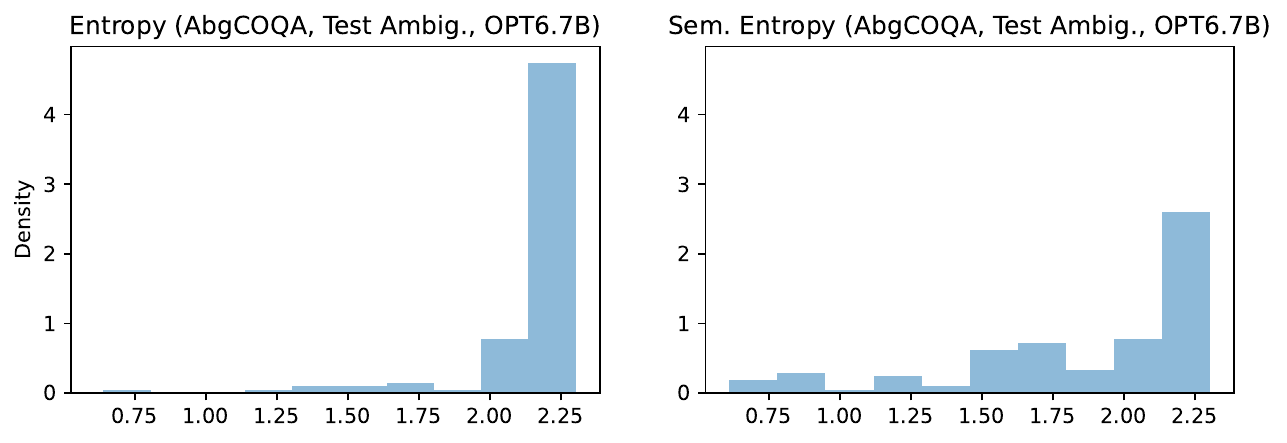}
    \hfill
    \includegraphics[width=6.5cm]{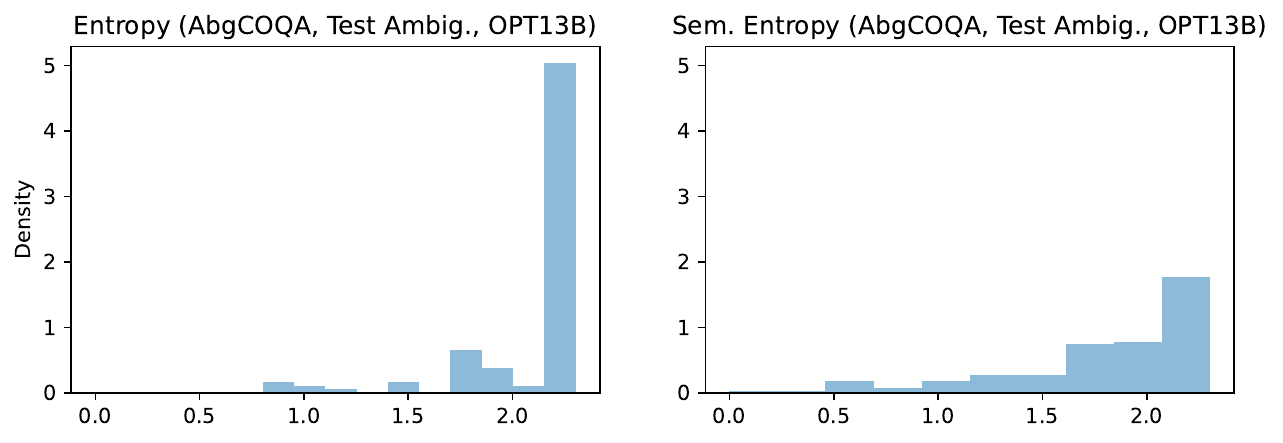}
    \hfill
    \includegraphics[width=6.5cm]{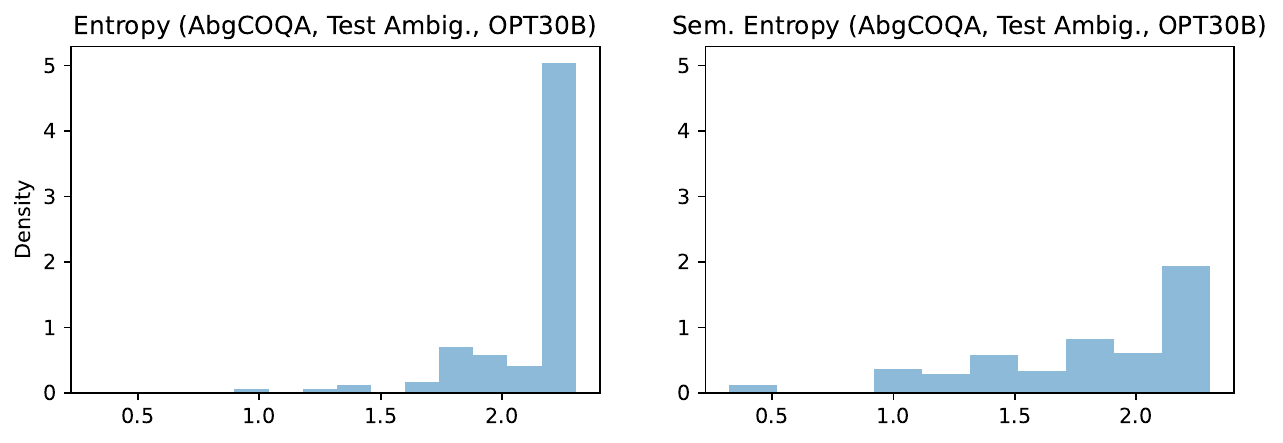}
    \hfill
    \caption{E and SE histograms across AbgCOQA's ambiguous prompts (from the test set), for different models.}
    \label{fig:e_se_abgcoqa}
\end{subfigure}
\centering
\begin{subfigure}[t]{6.5cm}
    \includegraphics[width=6.5cm]{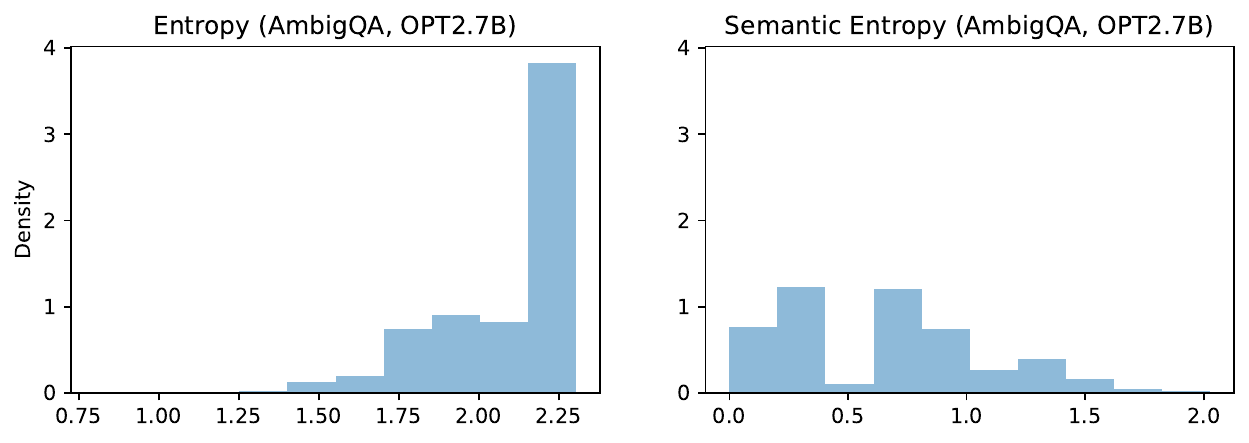}
    \hfill
    \includegraphics[width=6.5cm]{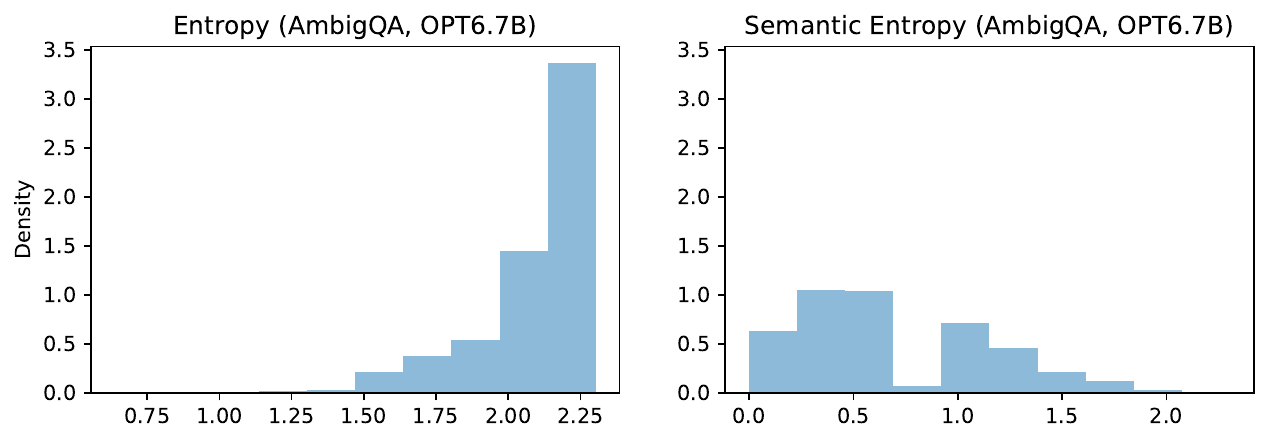}
    \hfill
    \includegraphics[width=6.5cm]{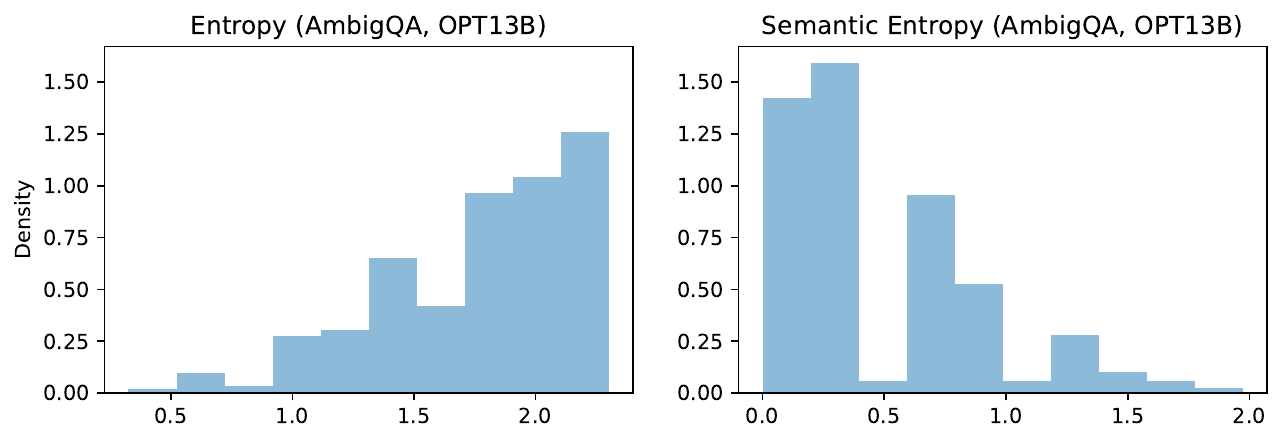}
    \hfill
    \includegraphics[width=6.5cm]{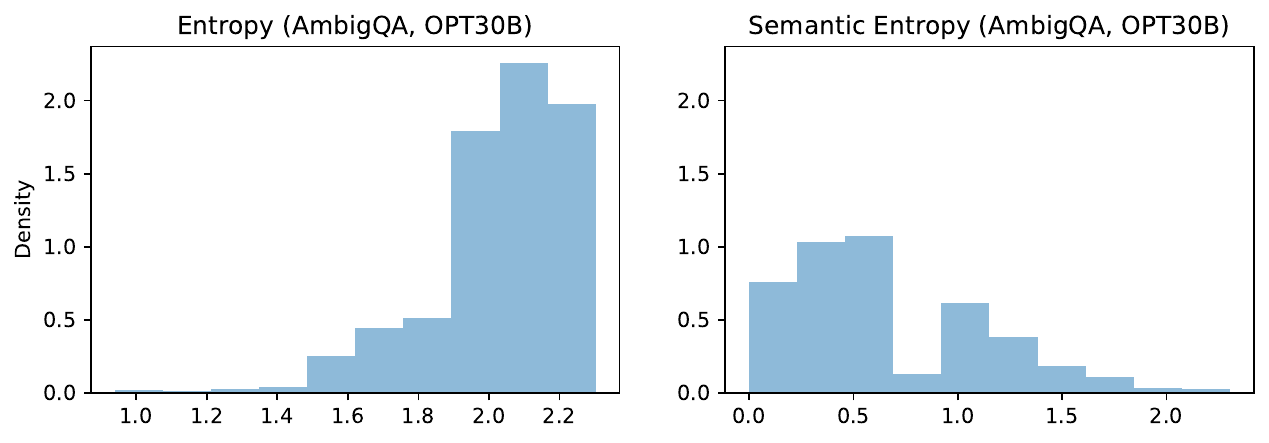}
    \hfill
    \caption{E and SE histograms across AmbigQA's prompts, for different models.}
    \label{fig:e_se_ambigqa}
\end{subfigure}
\caption{Entropy (E) and Semantic Entropy (SE), non-normalised for two datasets of ambiguous prompts.}
\label{fig:e_se_abgcoqa_ambigqa}
\end{figure*}

\begin{figure}[h!]
\includegraphics[width=6.5cm]{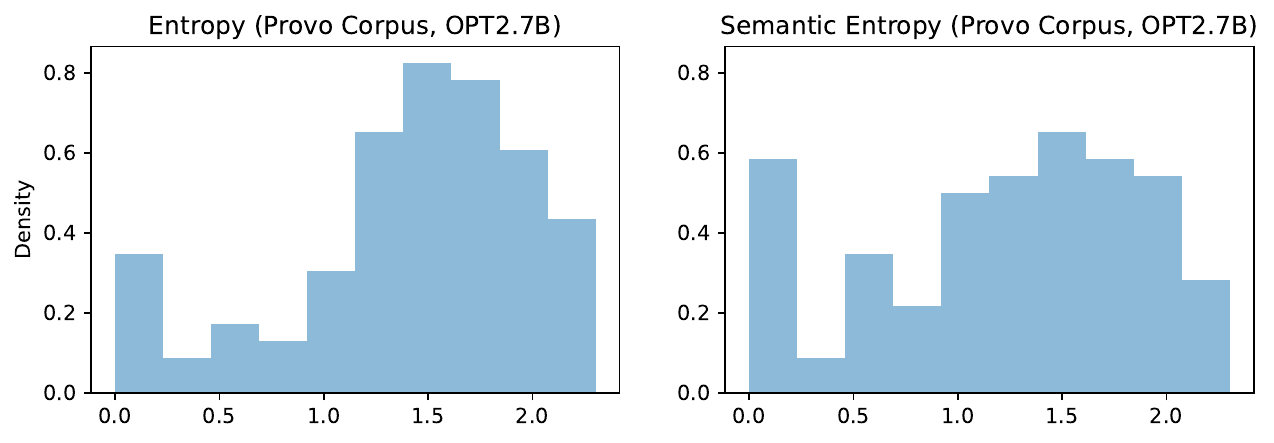}
\hfill
\includegraphics[width=6.5cm]{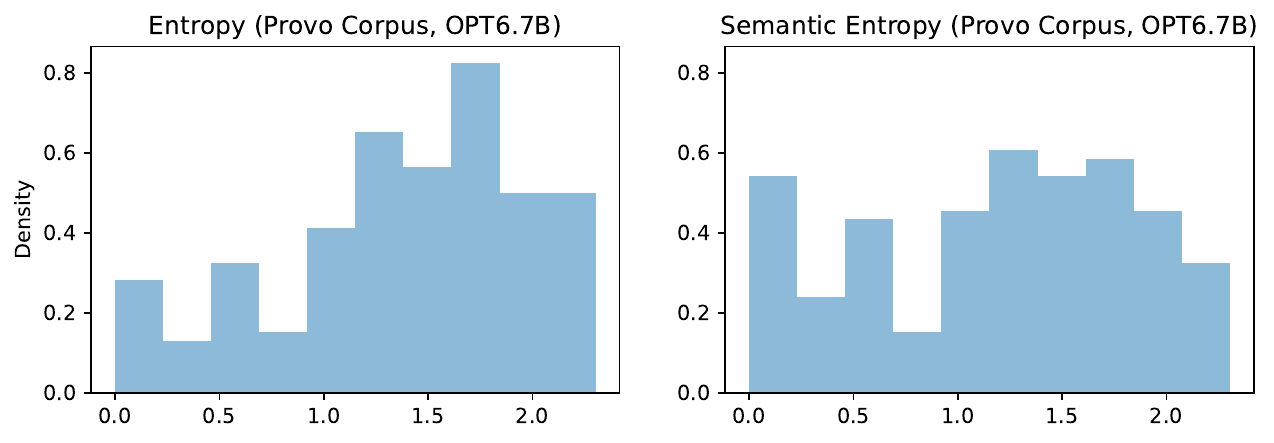}
\hfill
\includegraphics[width=6.5cm]{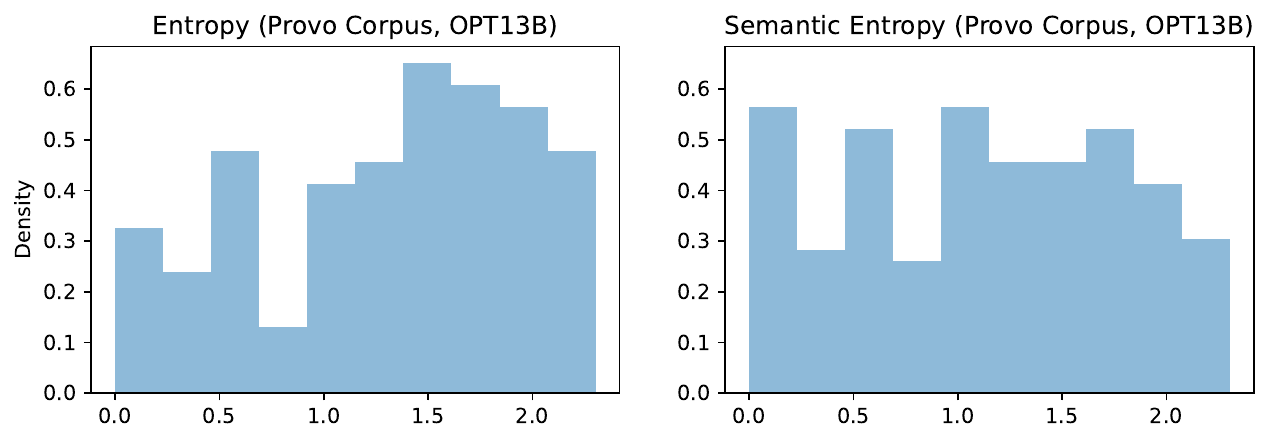}
\hfill
\includegraphics[width=6.5cm]{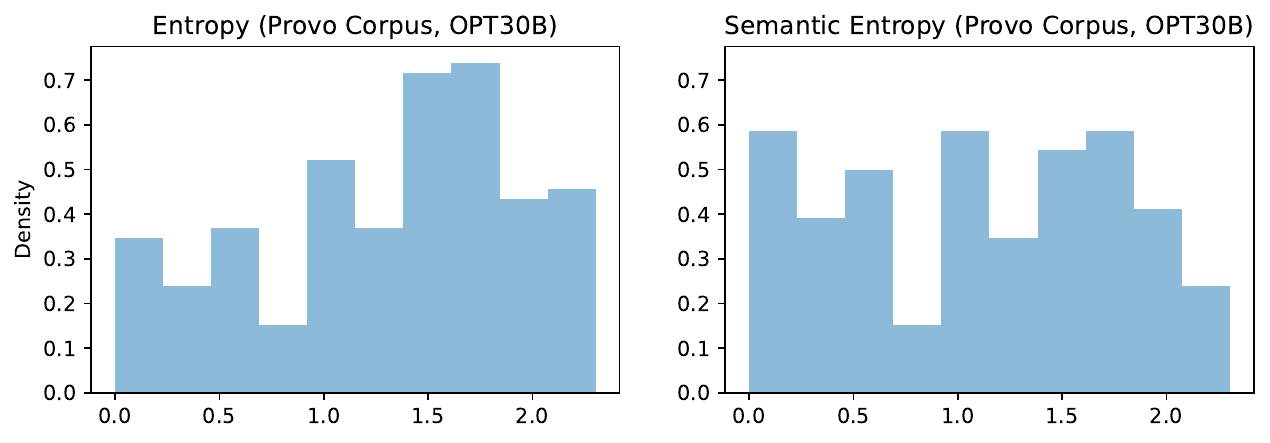}
\hfill
\caption{Entropy and Semantic Entropy histograms across Provo Corpus prompts, for different models.}
\label{fig:e_se_provo}
\end{figure}

Additionally, to get a grasp of the plausible variability we can expect for each dataset, we plot E and SE values (non-normalised) across prompts, for different models. 
In \Cref{fig:e_se_ambigqa}, we observe how despite variation among unique responses, the variation is among paraphrases (as the histograms of E and SE reveal). This is in contrast to variation observed in AbgCOQA and Provo Corpus, where variation among semantically distinct responses seems to exist (\Cref{fig:e_se_abgcoqa} and \ref{fig:e_se_provo}).

\section{Ablations}

\subsection{Sampled instead of greedy response for AUROC computation}
\label{appendix:sample_for_auroc}
\begin{figure}[t]
    \begin{center}
        \includegraphics[width=7 cm]{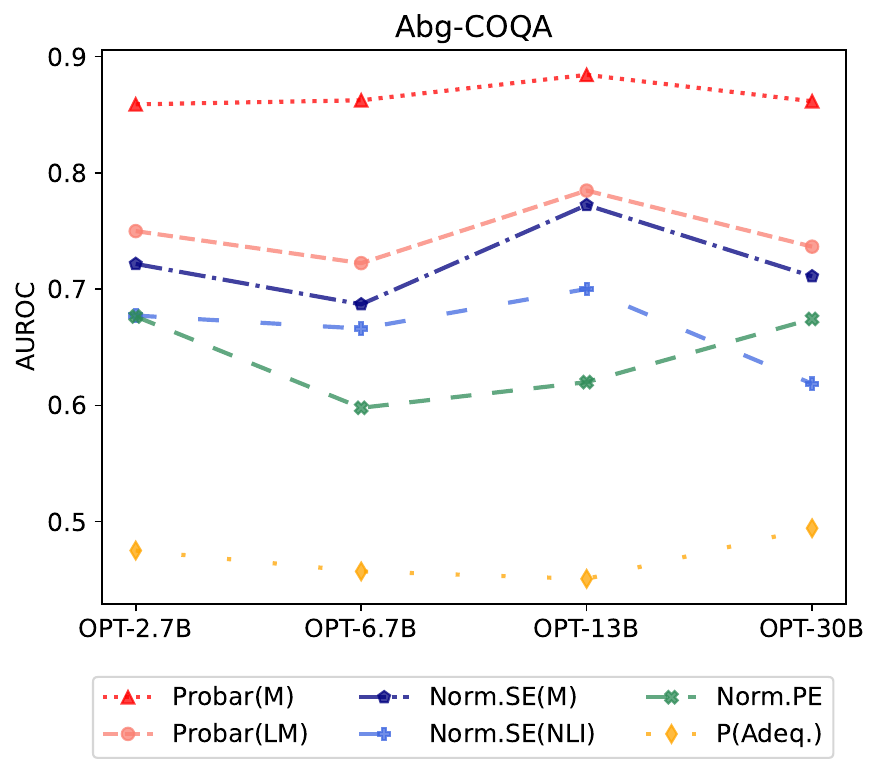}
        \caption{Average AUROC values (over 5 sampled responses) for various uncertainty indicators for the models analysed. They regard the 50 contexts that were manually annotated, and for the correctness of the sample, the manual annotation is used.}  
        \label{fig:auroc_vs_models_correct_sample}
    \end{center}
\end{figure}

We assessed whether SE and \probar are useful metrics for informing a user whether the greedy response is likely to be correct. We aim to assess whether similar trends hold when assessing how useful our metrics are when informing the user about the correctness of a sampled response (unbiasedely). We repeat our analysis on the 50 contexts for which we obtained the manual annotations and report the average AUROC over 5 runs (\emph{i.e.} over 5 different sampled responses, for which we sample from the already 10 sampled responses per context) for various uncertainty metrics. The results can be observed in \Cref{fig:auroc_vs_models_correct_sample}; the AUROCs shown are an average over 5 runs, while for the correctness of the sample we use the manual adequacy assessments we have annotated). We observe a similar trend to when the greedy response is used for the AUROC computation.

\subsection{Number of samples for \probar}
\label{appendix:num_samples_ablation}

\begin{figure}[t]
    \begin{center}
        \includegraphics[width=7 cm]{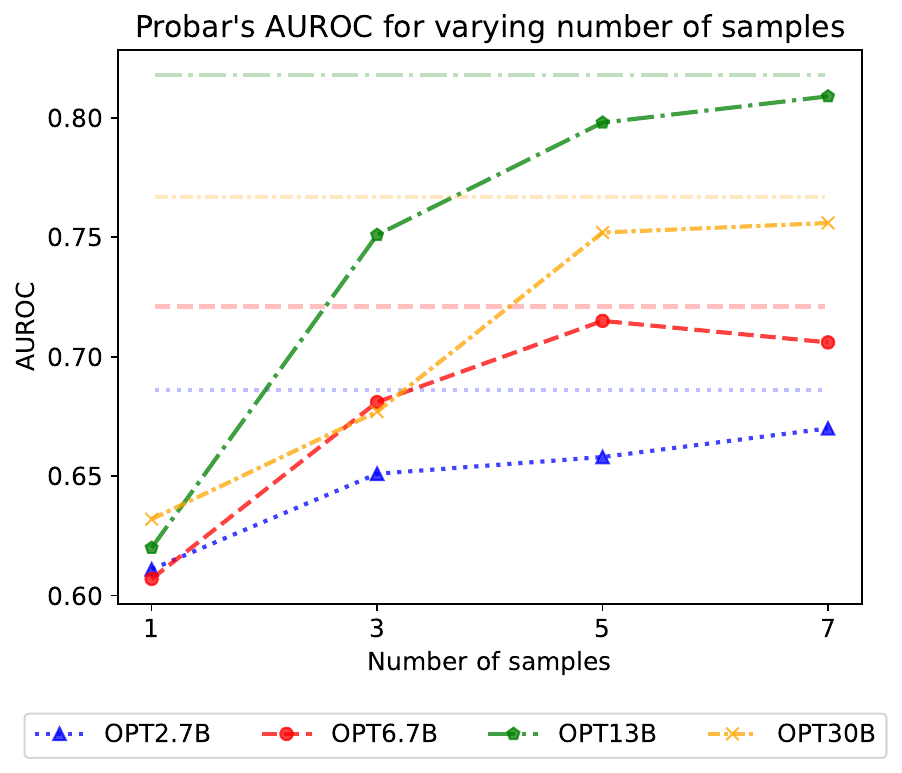}
        \caption{Probar's AUROC vs sample size (average of 5 runs). The faded-out lines show the AUROC values for the original 10 samples.}  
        \label{fig:probarauroc_vs_numsize}
    \end{center}
\end{figure}

Sampling-based confidence methods can be expensive. In the case of Probar, the quality of the classifier's predictions can obviously affect the informativeness of the metric overall. We speculate that a higher number of samples, will make \probar's performance more robust to the misclassifications of the adequacy classifier it relies on. We conduct a controlled experiment, where we subsample k samples from the 10 samples available to us for each context and compute \probar, for the 50 manually annotated contexts. The results can be seen in \Cref{fig:probarauroc_vs_numsize}: we observe how the performance of \probar improves as the number of samples grow. However, we can achieve reasonably similar performance to the original 10 samples using 5 samples or more.

\subsection{AUROC variance}
\label{appendix:ensemble_analysis}
\begin{figure}
\centering
\includegraphics[trim={0 0 1.95cm 0},clip, width=7.5cm]{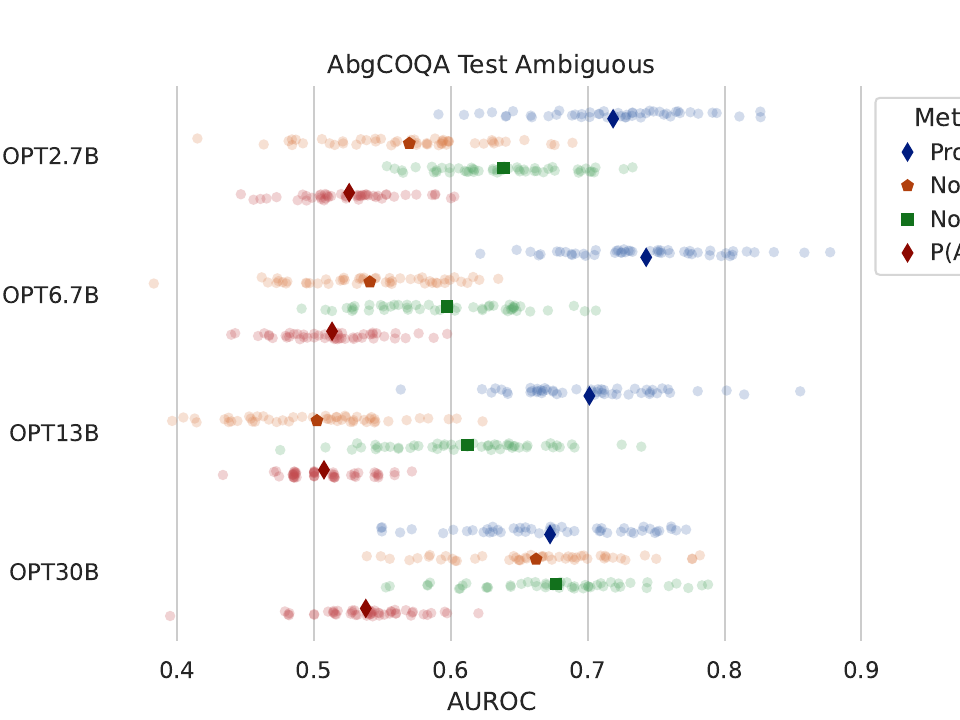}
\hfill
\includegraphics[trim={0 0 1.45cm 0},clip,width=7.5cm]{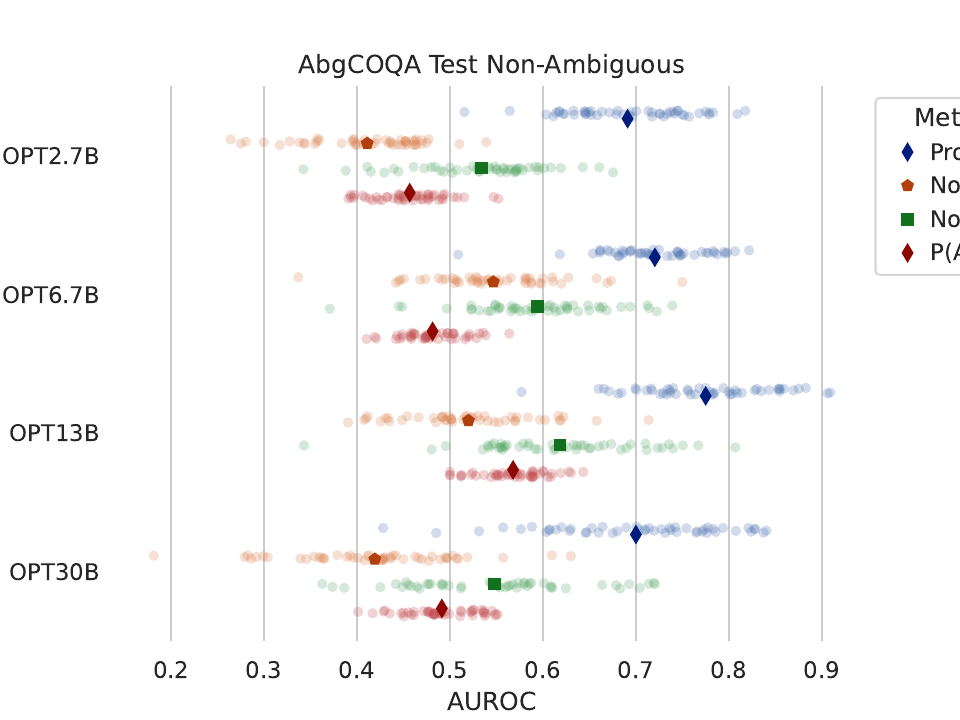}
\hfill
\includegraphics[trim={0 0 1.5cm 0},clip,width=7.5cm]{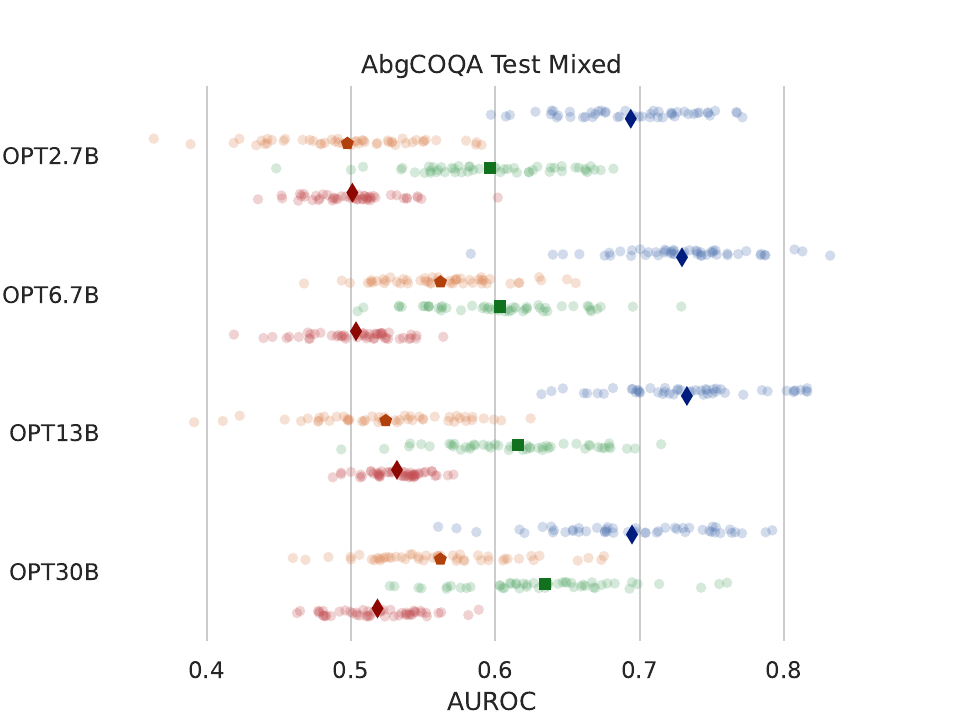}
\hfill
\includegraphics[trim={3cm 0 1cm 13cm},clip,width=6.5cm]{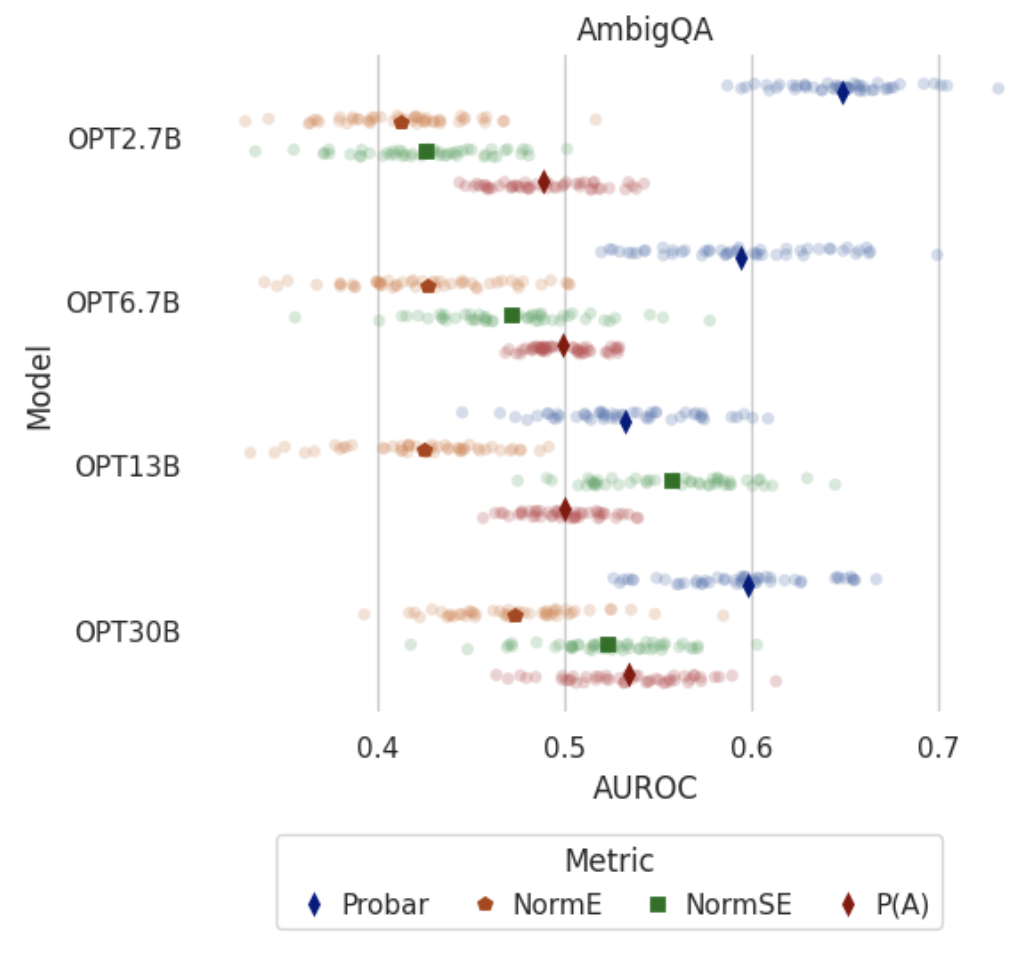}
\caption{AUROC values (and their mean) for 50 subsets of 50, 50, 100 (from top to bottom) for various models and quantifiers, for AbgCOQA's test set.}
\label{fig:variance}
\end{figure}

We assess how \probar 's performance varies (in terms of AUROC), depending on which prompts of the assessed set of prompts was used. We repeatedly subsample a subset of the contexts and compute AUROCs for SE, \probar and other quantifiers, which we present in \Cref{fig:variance}.

\begin{figure}
\centering
\includegraphics[trim={0 0 1.5cm 0},clip, width=7.5cm]{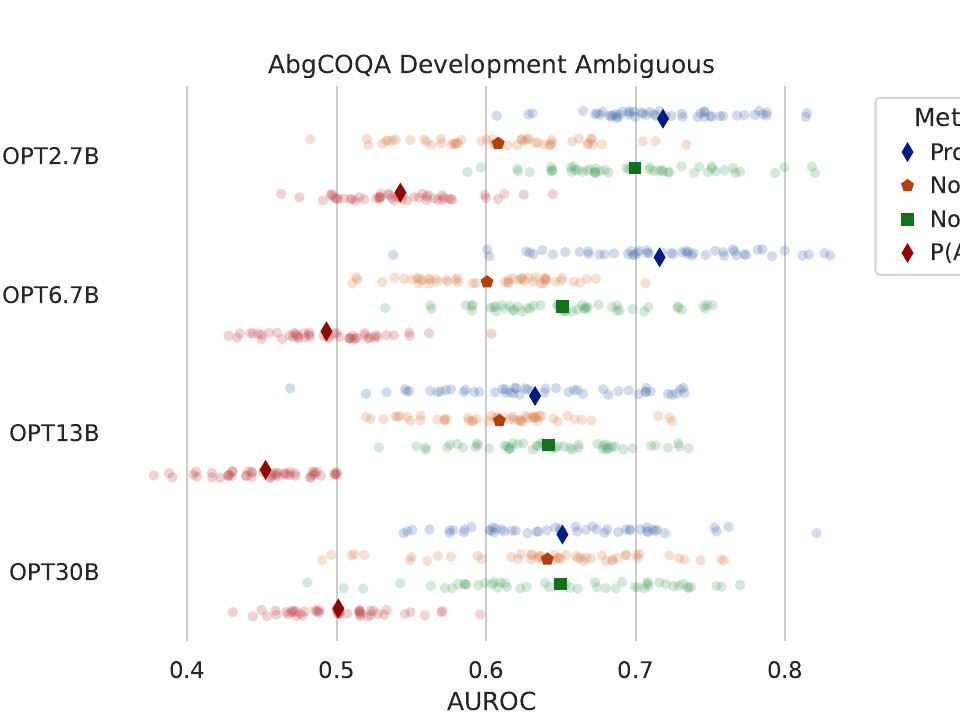}
\hfill
\includegraphics[trim={0 0 1.6cm 0},clip,width=7.5cm]{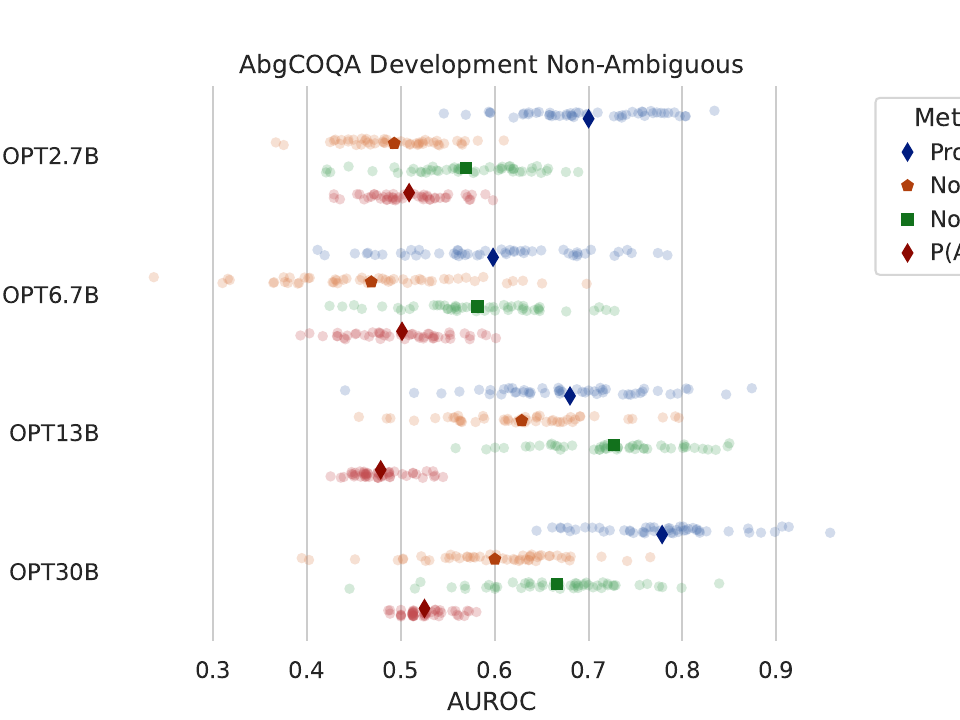}
\hfill
\includegraphics[trim={0 0 1cm 0},clip,width=7.5cm]{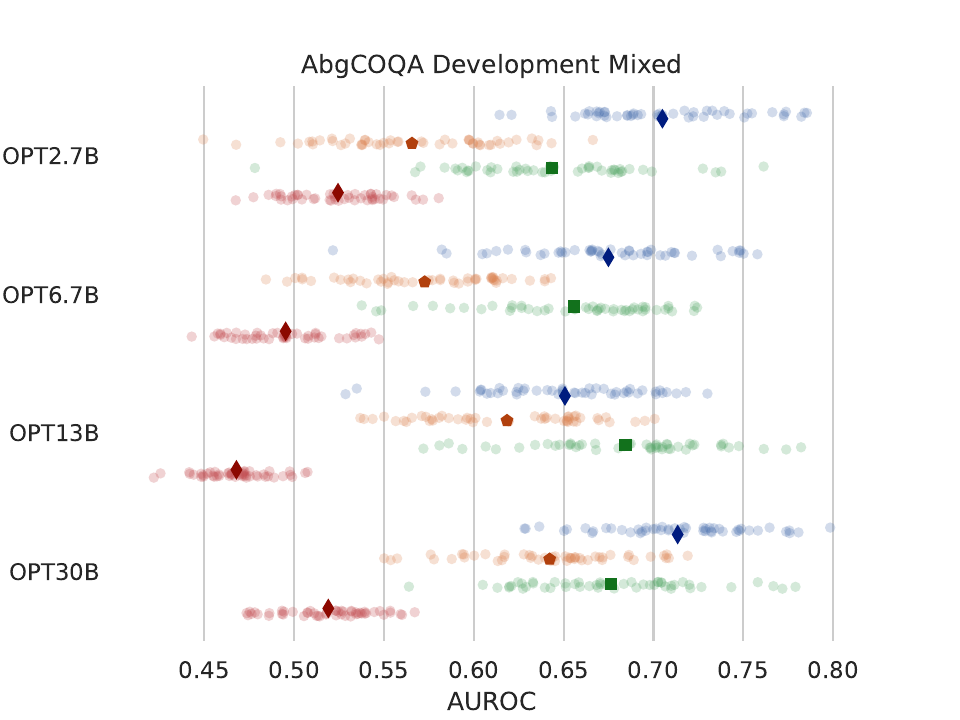}
\hfill
\includegraphics[trim={3cm 0 1cm 13cm},clip,width=6.5cm]{visuals/Screenshot.png}
\hfill
\caption{AUROC values (and their mean) for 50 subsets of 50, 50 and 100 (from top to bottom) for various models and quantifiers.}
\label{fig:variance_dev}
\end{figure}

\begin{figure}
    \centering
    \includegraphics[width=6.75cm]{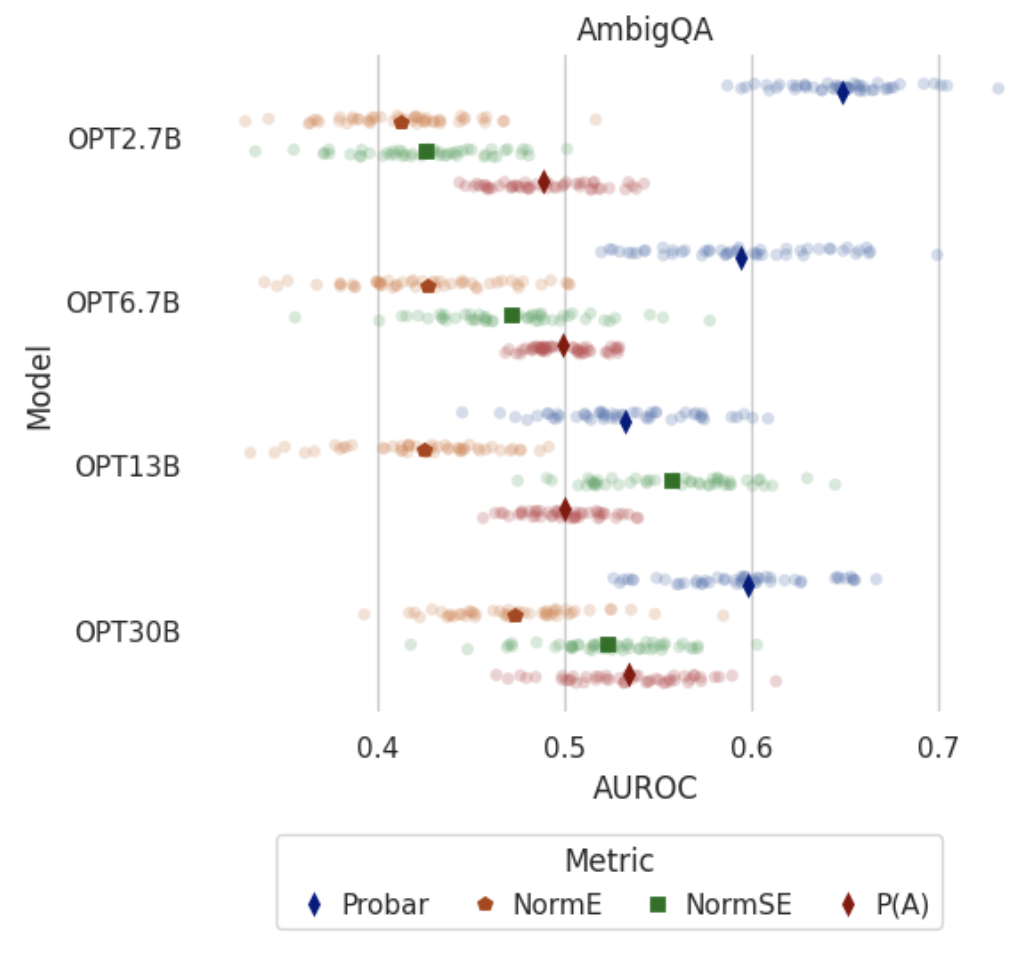}
    \caption{AUROC values (and their mean) for 50 subsets of 500 prompts for various models and quantifiers, for AmbigQA.}
    \label{fig:variance_ambigqa}
\end{figure}

\subsection{\probar for decoding}
\label{appendix:probar_decoding}
In our thus far analysis, \probar was employed as a metric that can inform the users about the model's reliability when responding to a prompt. We consider whether it could also be exploited for `confidence-infromed' decoding. As such, we conduct an experiment (using the subset of contexts we have manually annotated). For these, we assess precision of the greedy decoding (number of greedy decodings assessed as correct by manual annotations divided by number of greedy decodings). Regarding the \probar-aware decoding, for each prompt, we randomly sample one of the responses deemed as adequate by \probar's automated classifier and assess its manual correctness. If no samples were deemed as adequate by \probar's classifier, we abstain from generating. Precision is assessed again (number of sampled decodings assessed as correct by manual annotations divided by number of sampled decodings). Results can be seen in \Cref{tab:decoding_precision}; for the \probar -aware decoding performance, precision is an average over 10 runs (as there is randomness in the process). Although in one setting precision worsened (OPT13B), we observe some encouraging results, that could pave the path for future research in this direction.

\begin{table}[]
\begin{tabular}{l|rr}
 & \multicolumn{2}{l}{\textbf{Precision on decoding}} \\
 & \multicolumn{1}{l}{\textbf{Greedy}} & \multicolumn{1}{l}{\textbf{Probar-aware}} \\ \hline
\textbf{OPT2.7B} & 0.60 & \textbf{0.63} \\
\textbf{OPT6.7B} & \textbf{0.72} & \textbf{0.72} \\
\textbf{OPT13B} & \textbf{0.74} & 0.67 \\
\textbf{OPT30B} & 0.68 & \textbf{0.74}
\end{tabular}
\caption{Precision of Greedy vs Probar-aware decodings for prompts from different models, on the manually annotated subset of contexts. For Probar-aware results, the precision over 10 runs is shown.}
\label{tab:decoding_precision}
\end{table}

\end{document}